\documentclass{article}
\usepackage{float}
\usepackage{enumitem}
\usepackage{multirow}
\usepackage[utf8]{inputenc}
\usepackage[T1]{fontenc}
\usepackage{hyperref}
\usepackage{url}
\usepackage{amsfonts}
\usepackage{amsmath}
\usepackage{amssymb}
\usepackage{nicefrac}
\usepackage[dvipsnames]{xcolor}
\usepackage{microtype}
\usepackage{graphicx}
\usepackage{subcaption}
\usepackage{booktabs} % for professional tables

\usepackage{hyperref}

\usepackage[preprint]{icml2026}

\usepackage{amsmath}
\usepackage{amssymb}
\usepackage{mathtools}
\usepackage{amsthm}

\usepackage[table]{xcolor}
\usepackage{multirow}
\usepackage{makecell}
\usepackage{pifont}       
\usepackage{bbding}      
\usepackage{fontawesome}
\usepackage{subcaption}
\usepackage{arydshln}
\usepackage{tabularx}
\usepackage{mdframed}
\usepackage[most]{tcolorbox}
\usepackage{arydshln}
\usepackage{adjustbox}
\usepackage{placeins}
\usepackage[most]{tcolorbox}

\tcbuselibrary{breakable}
\usepackage{tcolorbox}
\usepackage{fvextra}

\definecolor{table-blue}{RGB}{173, 216, 230}
\definecolor{row-highlight}{RGB}{220, 230, 242}
\definecolor{qwen-header}{RGB}{235, 240, 248}
\definecolor{llama-header}{RGB}{248, 244, 235}
\definecolor{section-gray}{RGB}{245, 245, 245}
\definecolor{group-header}{RGB}{230, 235, 245}

\definecolor{zhz_gray}{rgb}{0.8,0.8,0.8}
\definecolor{darkgreen}{rgb}{0.0, 0.5, 0.0} % Custom dark green color
\definecolor{darkred}{rgb}{0.5, 0.0, 0.0}   % Custom dark red color
\usepackage[capitalize,noabbrev]{cleveref}

\theoremstyle{plain}

\theoremstyle{definition}

\theoremstyle{remark}

\usepackage[textsize=tiny]{todonotes}

\icmltitlerunning{Is MoE Routing a Huffman Code? Discovering the Frequency-Diversity Law in Chain-of-Thought}

\begin{document}
\onecolumn
  \icmltitle{Is MoE Routing a Huffman Code?\\Discovering the Frequency-Diversity Law in Chain-of-Thought}

  \begin{icmlauthorlist}
    \icmlauthor{Ching-Chieh Tsao}{ntu}
    \icmlauthor{Zhuoyi Lin}{astar}
    \icmlauthor{Wenya Wang}{ntu}
  \end{icmlauthorlist}

  \icmlaffiliation{ntu}{Nanyang Technological University}
  \icmlaffiliation{astar}{Institute for Infocomm Research (I2R), A*STAR}

  \icmlcorrespondingauthor{Wenya Wang}{wangwy@ntu.edu.sg}
  % You may provide any keywords that you find helpful for describing your
  % paper; these are used to populate the "keywords" metadata in the PDF but
  % will not be shown in the document
  \icmlkeywords{Mixture-of-Expert, Model Reasoning}

  \vskip 0.3in

% this must go after the closing bracket ] following \twocolumn[ ...

% This command actually creates the footnote in the first column listing the
% affiliations and the copyright notice. The command takes one argument, which
% is text to display at the start of the footnote. The \icmlEqualContribution
% command is standard text for equal contribution. Remove it (just {}) if you
% do not need this facility.

% Use ONE of the following lines. DO NOT remove the command.
% If you have no special notice, KEEP empty braces:
% \printAffiliationsAndNotice{}  % no special notice (required even if empty)
% Or, if applicable, use the standard equal contribution text:
\printAffiliationsAndNotice{}

\begin{abstract}

Mixture-of-Experts architectures have revolutionized scaling, yet the underlying logic of their routing remains a black box. In this paper, we uncover a fundamental governing principle: \emph{MoE routing is not merely selection, but a manifestation of Huffman Coding.} We introduce the \textbf{Frequency-Diversity Law}, revealing that state-of-the-art models, such as Phi-3.5-MoE, Gemma-4-27B-A4B, spontaneously act as information-theoretic engines. These models allocate sparse expert resources for common tokens while invoking high-diversity expert committees for rare, complex tasks found in chain-of-thought trajectories. However, we identify a critical redundancy trap in Qwen3.5-35B-A3B: when effective sparsity ($k/E_{\text{eff}}$) is sufficiently low, load-balancing inadvertently imposes functional redundancy, masking the underlying Huffman efficiency signal. To bridge this gap, we propose \textbf{Subset Difference Pruning}, a surgical strategy to eliminate functional duplicates. We demonstrate that pruning does not degrade reasoning; instead, it unleashes the model’s latent Huffman efficiency, forcing the logic to collapse into streamlined, high-density paths. Our findings suggest that the next generation of MoEs should move beyond forced load-balancing toward \textbf{Minimum Description Length (MDL) optimality}, assigning shorter expert-routing codes to high-frequency information and longer, more diverse codes to low-frequency information, thereby transforming routing from a heuristic into a principled compression engine.

\end{abstract}

\section{Introduction}
\label{sec:intro}

Mixture-of-Experts (MoE) architectures have emerged as a dominant paradigm for scaling Large Language Models (LLMs) \citep{shazeer2017outrageously, jiang2024mixtral, dai2024deepseekmoe}, primarily due to their ability to decouple model capacity from computational cost. At the heart of this efficiency is the Gating Network (Router), which dynamically selects a small subset of experts to process each input token \citep{fedus2022switch}. Fundamentally, this process is an exercise in \textbf{resource allocation} \citep{clark2022unified}: the model must decide, in real-time, how to distribute its limited \emph{active} parameters to best represent the vast complexity of world knowledge.

This mechanism of selective activation mirrors the core principles of \textbf{Information Theory}, specifically \textbf{Source Coding} \citep{Shannon1948method}. Huffman Coding \citep{huffman1952method}, one of the cornerstone of this field, serves as the benchmark for data compression and communication efficiency. Its essence lies in allocating resources based on symbol probability: assigning shorter codes to high-frequency information and longer codes to low-frequency information. If an LLM is viewed as a predictive system aimed at the ``ultimate compression'' of world knowledge \citep{hutter2005universal}, its MoE reasoning trajectory should not be random; rather, it should reflect the frequency distribution of semantic logic, achieving an optimal balance between information entropy and physical resources. \citep{shannon1959coding, shazeer2017outrageously}

Despite this natural connection, the potential alignment between MoE routing and Huffman coding principles remains relatively under-explored. While existing literature has extensively examined scaling laws \citep{krajewski2024scaling}, expert specialization \citep{zhou2022mixture, xue2024openmoe}, and routing stability \citep{lepikhin2021gshard}, an information-theoretic characterization of routing efficiency from the perspective of source coding has yet to be fully established.

\begin{figure}[t]
\centering
\small \textbf{Huffman test: $\bar{u}^{(o)} \propto -\log p(o)$} \\ [1.0ex]
\includegraphics[width=0.75\textwidth]{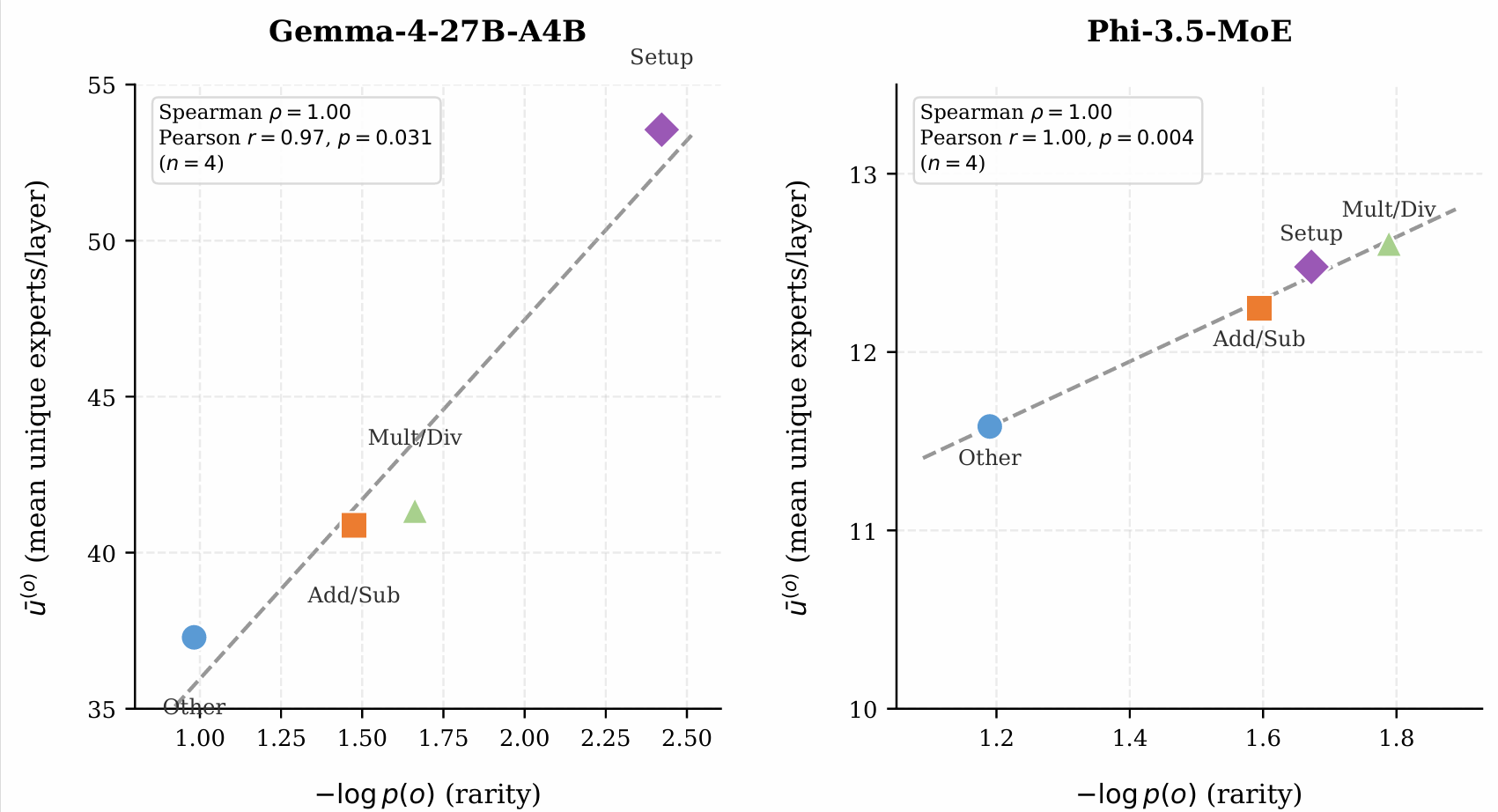}
\caption{%
    \textbf{Quantitative test of the Frequency-Diversity Law (Eq.~\eqref{eq:huffman}) on Gemma-4-27B-A4B and Phi-3.5-MoE.}
    Each point is one operation type $o \in \mathcal{O}$;
    $x$-axis: rarity $({-}\log p(o))$; $y$-axis: mean unique experts per layer ($\bar{u}^{(o)}$); dashed line: OLS fit.
    Both models achieve Spearman $\rho = 1.00$ ($p < 0.005$), confirming $\bar{u}^{(o)} \propto {-}\log p(o)$.}
\label{fig:huffman_scatter}
\end{figure}

To verify our hypothesis that optimized MoEs naturally gravitate towards Huffman-like efficiency, we conducted quantitative experiments on state-of-the-art models such as Gemma-4-27B-A4B and Phi-3.5-MoE. As illustrated in Figure \ref{fig:huffman_scatter}, we discovered a significant \emph{\textbf{Huffman Phenomenon}}: the number of activated experts (mean unique experts/layer) exhibits an extremely strong positive correlation with task rarity ($-\log p(o)$), with a Spearman's $\rho = 1.00$. This confirms that driven by the minimization of prediction loss, these models have automatically optimized routing efficiency by utilizing a few core experts for general logic and invoking a more diverse combination of experts only when encountering rare and complex tasks.

Nevertheless, the determination of expert numbers in current MoE designs remains largely empirical \citep{zoph2022designing, krajewski2024scaling}. Modern training paradigms typically prioritize suppressing ``expert collapse'' by introducing auxiliary load-balancing loss to ensure uniform communication across the expert pool \citep{yang2025qwen3, Kamath2025Gemma3T, fedus2022switch, omi2025load}. Our empirical analysis of Qwen3.5-35B-A3B reveals that when expert sparsity ($k/E_{\text{eff}}$) is excessively small, these balancing pressures may inadvertently compel the model to overallocate gradients to underutilized parameters (Table~\ref{tab:qwen_sim}). This manifests as a proliferation of \emph{functional clones} \citep{puigcerver2023sparse}, which are redundant experts with high semantic overlap that exist primarily to satisfy balancing objectives rather than to provide unique informational gain.

This leads to a pivotal inquiry: \emph{\textbf{Is the conventional reliance on aggressive load-balancing an optimal scaling strategy, or does it fundamentally obstruct the emergence of Huffman-efficient routing?}} We argue that routing essentially promotes compression: \textbf{the expert activation pattern selected by a router functions as a compact code for the semantic operations unfolding in a CoT \citep{wei2022chain} trace}. Motivated by this perspective, our investigation reveals that while models like Qwen3.5-35B-A3B maintain high active expert counts, the resulting code is often inefficient due to functional redundancy.

To test whether removing such redundancy can unleash latent Huffman efficiency, we introduce \textbf{Subset Difference Pruning} (SDP), a surgical strategy designed to refine the model’s routing topology. Our findings indicate that strategically removing moderately redundant experts effectively reshapes the model's reasoning path. Experimental results demonstrate that despite trajectory shifts caused by pruning, the model exhibits a robust re-encoding capability, collapsing logic into a more streamlined efficient path that aligns better with Huffman principles.

The contribution of this research lies in:
\begin{enumerate}[leftmargin=1.5em, itemsep=0pt, topsep=2pt]
    \item \textbf{Phenomenon:} We identify the \textit{Huffman encoding phenomenon} in MoE routing, revealing the temporal compression of expert patterns during CoT.
    \item \textbf{Universality:} We establish the \textit{Frequency-Diversity Law}, proving that frequency-diversity correspondence in expert routing is a universal structural property of optimized MoE models rather than a model-specific coincidence.
    \item \textbf{Methodology:} We propose \textit{Subset Difference Pruning} and identify the effective sparsity ($k/E_{\text{eff}}$) range for routing to operate as a genuine MDL-optimal compression engine.
\end{enumerate}

\section{Related Work}
\label{sec:related}
\subsection{Sparse Mixture-of-Experts (MoE)}Sparse MoEs \citep{shazeer2017outrageously} utilize top-$k$ routing to achieve high parameter counts with constant computational costs. While early scaling focused on massive deployments \citep{lepikhin2021gshard, fedus2022switch}, recent architectures emphasize refined expert granularity. \citet{yang2025qwen3} and \citet{Kamath2025Gemma3T} introduced diverse expert configurations, such as shared-expert isolation, to optimize utilization. We analyze modern benchmarks, Phi-3.5-MoE \citep{abdin2024phi}, Gemma-4-27B-A4B \citep{Kamath2025Gemma3T}, and Qwen3.5-35B-A3B \citep{yang2025qwen3}, which represent distinct balances between expert pool size and routing sparsity ($k/E \in \{3.1\%, 6.25\%, 12.5\%\}$), providing a spectrum to study information-theoretic routing properties.

\subsection{Routing Interpretability and Dynamics}
Interpretability research often treats MoE experts as specialized key-value memories \citep{geva2021transformer, elhage2022solu}. Previous studies explored whether experts encode semantic or syntactic features \citep{artetxe2022efficient, herbst2026expert} but largely relied on static, single-token snapshots. Recent systematic analyses have begun to uncover macro-structures in routing, such as layer-wise specialization and ``Language Routing Isolation,'' where experts are partitioned by linguistic resource levels and families \citep{chen2026understanding, zheng2026unveiling}. However, these insights remain focused on the spatial distribution of experts across model depth. Our work departs from this by analyzing the \textbf{temporal dynamics} of routing, specifically across CoT trajectories. We bridge a critical gap by linking routing decisions to Huffman coding, demonstrating that routers spontaneously optimize for frequency-dependent expert allocation when given adequate expert sparsity.

\subsection{Information-Theoretic Compression}The Minimum Description Length (MDL) principle \citep{rissanen1978modeling} provides a unified framework for weight pruning \citep{han2015learning} and quantization \citep{frantar2022gptq}. While LLMs are recognized as general-purpose compressors \citep{deletang2023language}, and concurrent work like RFID-MoE \citep{mi2026effective} uses routing frequency for weight-SVD compression, our approach is distinct. We focus on \textbf{routing path description length} rather than parameter weights. By drawing on optimal source coding \citep{cover2006elements}, we argue that inefficient models (e.g., Qwen3.5-35B-A3B) suffer from functional redundancy where forced load-balancing imposes a sub-optimal prior on the routing distribution. Our Subset Difference Pruning targets this ``routing codebook'' redundancy, recovering MDL-optimality without further training.

\section{Temporal Expert Combination Encoding: Information-Theoretic Analysis}
\label{sec:huffman}
This section formalizes the claim that MoE routing implements Huffman coding. We define expert activation patterns as combinatorial codewords, connect them to information theory via mutual information, and derive the \emph{Frequency-Diversity Law} as a falsifiable prediction. We then show this law extends temporally along individual CoT traces. Throughout, we adopt the thought-step segmentation convention from \citet{zhou2025landscape}, treating each segment of a CoT trace demarcated by newlines as a distinct reasoning step.

\paragraph{Expert Set and Unique Expert Count.}
For each reasoning step $i$ in a CoT trace, define the \emph{expert set} at layer $l$
as the union of all activated expert indices across the tokens of that step:
\begin{equation}
    \mathcal{E}_i^{(l)} = \bigcup_{t \in \mathcal{T}_i} \operatorname{top\text{-}k}
    \!\left(\mathbf{g}_t^{(l)}\right),
\end{equation}
where $\mathbf{g}_t^{(l)}$ is the gating weight vector at token $t$ and layer $l$,
and $\mathcal{T}_i$ denotes the token indices belonging to step $i$.
The \emph{unique expert count} at step $i$ is denoted by $u_i = |\mathcal{E}_i^{(l)}|$, serving as the code-length surrogate in the analysis below.

\begin{figure}[t]
\centering
\small \textbf{Unique Expert Count Trajectory -- MATH by level} \\ [1.0ex]
\begin{subfigure}[b]{0.31\textwidth}
    \centering
    \includegraphics[width=\textwidth]{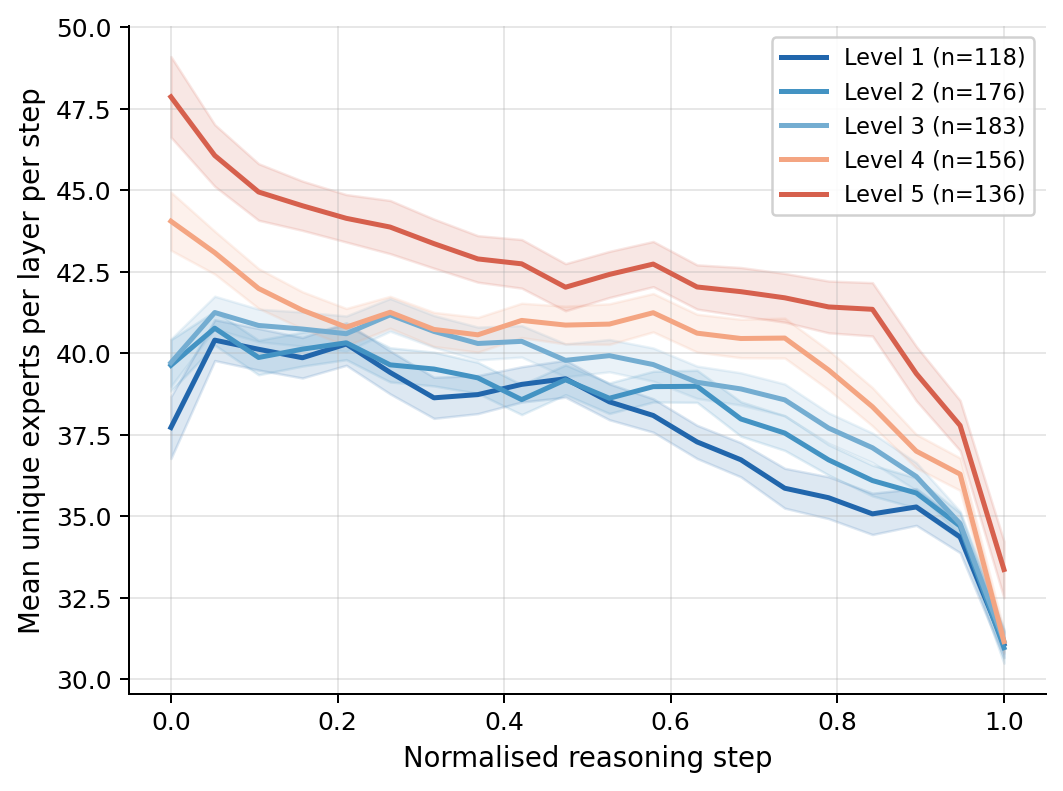}
    \caption{Gemma-4-27B-A4B}
    \label{fig:unique_expert:gemma}
\end{subfigure}
\hfill
\begin{subfigure}[b]{0.31\textwidth}
    \centering
    \includegraphics[width=\textwidth]{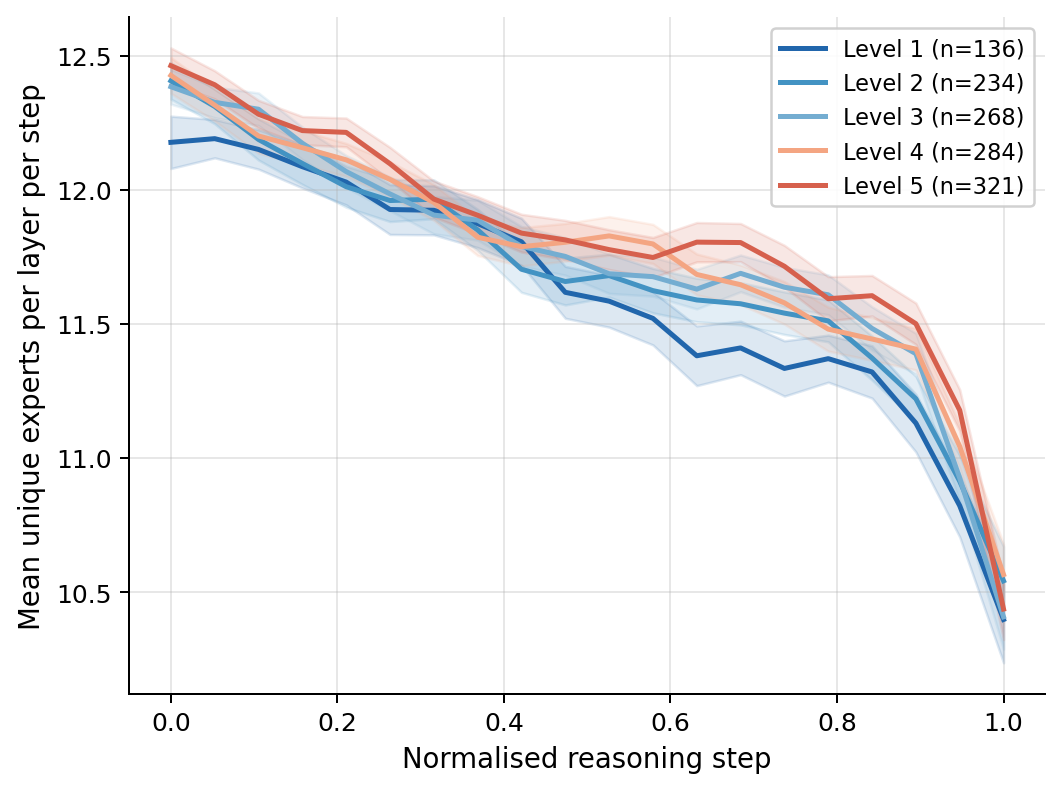}
    \caption{Phi-3.5-MoE}
    \label{fig:unique_expert:phi}
\end{subfigure}
\hfill
\begin{subfigure}[b]{0.31\textwidth}
    \centering
    \includegraphics[width=\textwidth]{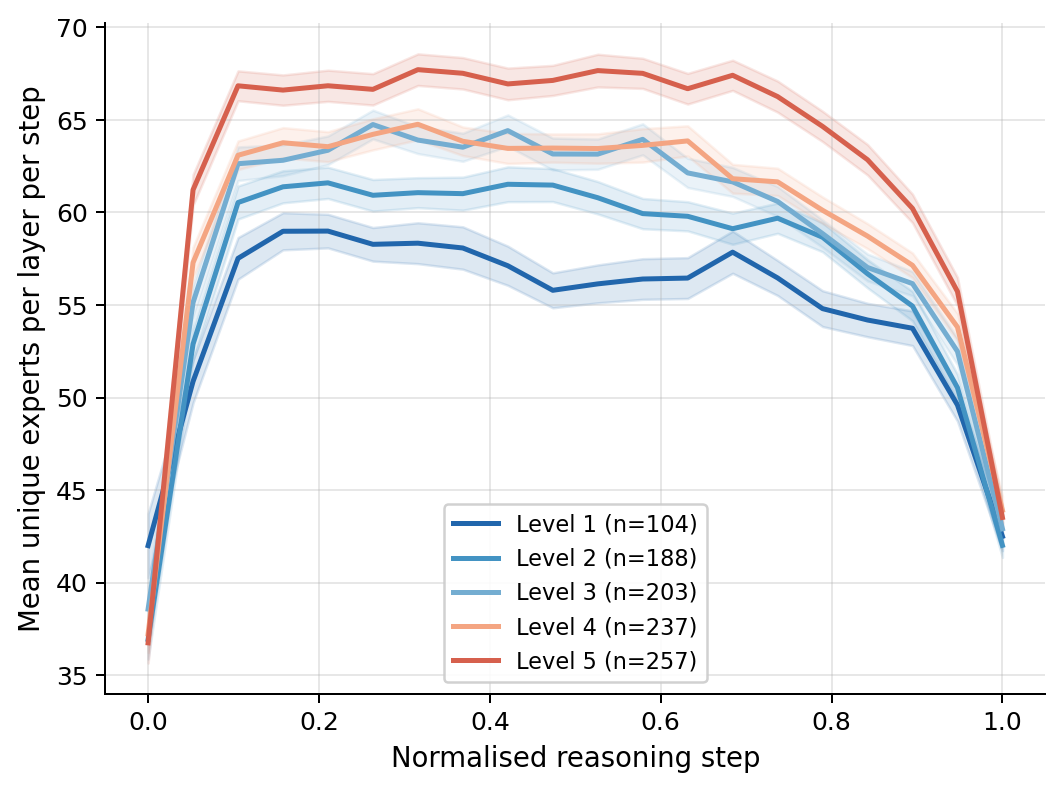}
    \caption{Qwen3.5-35B-A3B}
    \label{fig:unique_expert:original}
\end{subfigure}
\vspace{1ex}
\caption{Unique expert count trajectory across normalized reasoning steps (MATH) for Gemma-4-27B-A4B, Phi-3.5-MoE, and Qwen3.5-35B-A3B.}
\label{fig:unique_expert_comparison}
\end{figure}

\paragraph{Router as an Information-Theoretic Encoder.}
We interpret the MoE router as a \emph{semantic encoder}: the gating mechanism maps
the input at step $i$ to a discrete expert activation pattern $\mathcal{E}_i^{(l)}$,
which functions as a combinatorial codeword drawn from the expert index set
$\{1, \ldots, E\}$.

Let $O \in \mathcal{O}$ denote the \emph{semantic operation type} of a reasoning step, where
$\mathcal{O} = \{\texttt{add/subtract},\,\texttt{multiply/divide},\,\texttt{problem setup},\,\texttt{other}\}$ \footnote{These four categories are selected because LLMs are inherently limited in performing highly complex arithmetic computations; problem setup and the four basic arithmetic operations represent the core cognitive capabilities required for LLM Chain-of-Thought reasoning on mathematical problems.}
is classified per step using a lightweight keyword classifier (Appendix~\ref{app:specialization}),
and $p(o)$ denotes the empirical frequency of type $o \in \mathcal{O}$.
The quality of routing is measured by the \emph{mutual information} between $O$
and the induced activation pattern $\mathcal{E}^{(l)}$:
\begin{equation}
    I\!\left(O;\, \mathcal{E}^{(l)}\right)
    = H(O) - H\!\left(O \mid \mathcal{E}^{(l)}\right),
\end{equation}
where $H(O) = -\sum_{o \in \mathcal{O}} p(o)\log p(o)$ is the semantic entropy and
$H(O \mid \mathcal{E}^{(l)})$ is the residual uncertainty after observing the routing
pattern. A well-optimized router maximizes $I(O;\mathcal{E}^{(l)})$, ensuring the
activated expert set is as informative as possible about the underlying semantic operation.

\paragraph{From Mutual Information to Huffman Optimality.}
A key result from information theory \citep{cover2006elements} is that maximizing
mutual information under a mean activation constraint is equivalent to minimizing
the average description length of the source. Interpreting $u_i$ as the \emph{code
length} assigned to step $i$, the Huffman lower bound gives:
\begin{equation}
    \mathbb{E}[u_i]
    \;\geq\; H(O) / \log\binom{E}{k},
    \label{eq:entropy_bound}
\end{equation}
with equality when shorter codes (smaller $u_i$) are assigned to more frequent
operation types. This yields the \textbf{Frequency-Diversity Law} as a falsifiable
prediction:
\begin{equation}
    \bar{u}^{(o)} \;\propto\; {-}\log p(o), \label{eq:huffman}
\end{equation}
where $\bar{u}^{(o)} = \mathbb{E}[u_i \mid \operatorname{type}(i) = o]$ is the mean
unique expert count for steps of type $o \in \mathcal{O}$.
Equation~\eqref{eq:huffman} states that the router assigns shorter combinatorial codes
to more frequent operation types, satisfying the Huffman optimality condition that
maximizes $I(O;\mathcal{E}^{(l)})$ under the mean activation constraint.

\begin{table}[t]
\caption{Routing architecture characteristics and their effect on expert space \\pruning manifestation.}
\label{tab:routing_summary}
\centering
\small
\begin{tabular}{lcccl}
\toprule
Model & $E$ & $k$ & $k/E$ & Pruning Pattern \\
\midrule
Gemma-4-27B-A4B & 128 & 8 & 6.25\% & Monotonic decrease in $u_i$ \\
Phi-3.5-MoE     &  16 & 2 & 12.5\% & Oscillating (cluster pairs) \\
Qwen3.5-35B-A3B & 256 & 8 & 3.1\% & Sparsity masks signal (Huffman unmeasurable) \\
\bottomrule
\end{tabular}
\end{table}

\paragraph{Temporal Compression: Expert Diversity Decays Along the Reasoning Trace.}
Beyond the cross-type Huffman correspondence, we observe a complementary 
\emph{temporal} phenomenon within individual CoT traces. As illustrated in
Figure~\ref{fig:unique_expert_comparison}, the unique expert count $u_i$ typically peaks during the early reasoning phases before exhibiting a systematic contraction in the latter half of the trace as the index $i$ increases.

We interpret this as an information convergence process. In the early steps 
of a CoT trace, the reasoning state is highly uncertain: the model must explore a 
broad semantic space to parse the problem and identify relevant sub-tasks. This 
corresponds to high conditional entropy $H(O \mid \text{step } i)$, which, according to the 
Huffman correspondence in Eq.~\eqref{eq:huffman}, induces larger expert sets 
(longer codes). As reasoning progresses, the model incrementally resolves this 
uncertainty, concentrating probability mass onto a narrowing set of operation types. 
The resulting reduction in semantic entropy manifests as a contraction of the 
activated expert set, with later steps relying on a compact core of high-frequency 
experts. Formally, this temporal dynamic can be expressed as:
\begin{equation}
    \mathbb{E}[u_i] \;\approx\; \alpha \cdot H\!\left(O \mid \text{step } i\right) 
    + \beta, \quad \alpha > 0,
    \label{eq:temporal_decay}
\end{equation}
where $\beta$ is the baseline expert count when semantic uncertainty is fully resolved ($H = 0$), and $\alpha > 0$ quantifies the router's sensitivity: each additional nat of operation-type uncertainty induces $\alpha$ more unique experts per layer.
We estimate $\hat{H}(O \mid \text{step } i)$ empirically by partitioning the normalized trace into ten equal-width position bins and computing the entropy of the empirical operation-type distribution within each bin.

\begin{figure}[t]
    \centering
    \small \textbf{Temporal Compression: $\bar{u}_i$ vs.\ $\hat{H}(O \mid \text{step}\,i)$ --- GSM8K} \\ [1.0ex]
    \includegraphics[width=0.95\linewidth]{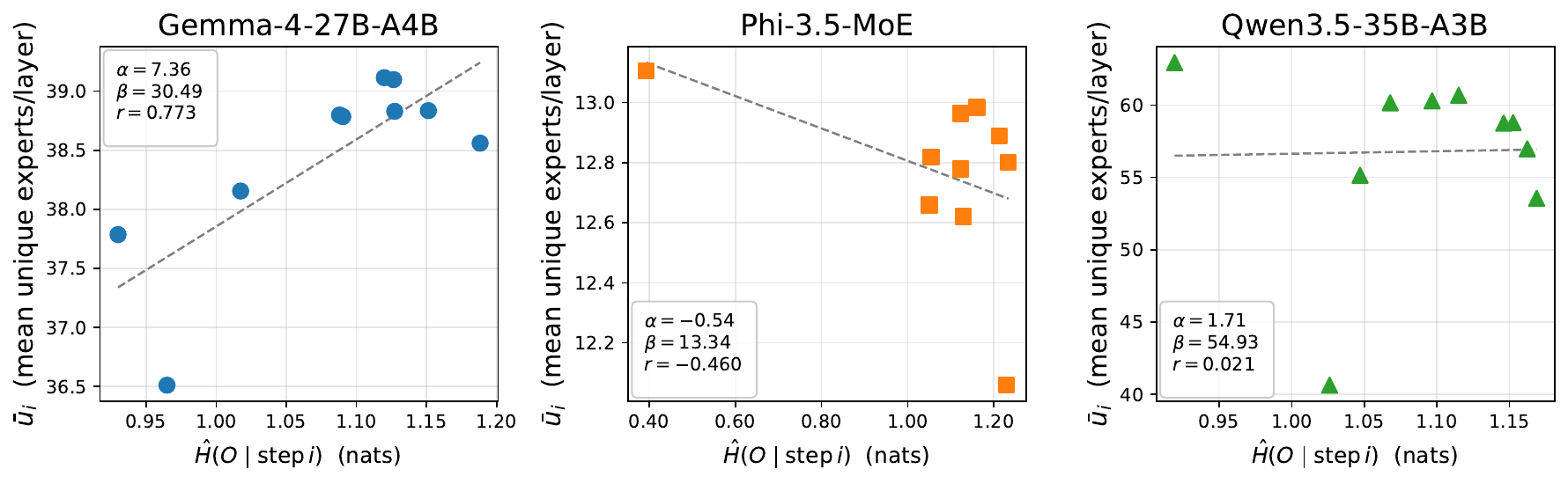}
    \caption{$\bar{u}_i$ vs.\ empirical $\hat{H}(O \mid \text{step}\,i)$ on GSM8K (ten decile bins per model; dashed line: OLS fit).
    Gemma confirms the linear relationship in Eq.~\eqref{eq:temporal_decay} ($\alpha{=}7.36$, $\beta{=}30.49$, $r{=}0.773$, $p{=}0.009$).
    Phi ($r{=}{-}0.460$, $p{=}0.18$) deviates because its oscillating trajectory makes $\hat{H}$ a poor linear predictor.
    Qwen ($r{\approx}0$) shows no relationship, consistent with functional redundancy decoupling expert count from operation-type entropy.}
    \label{fig:temporal_decay_fit}
\end{figure}

Figure~\ref{fig:temporal_decay_fit} validates Eq.~\eqref{eq:temporal_decay} on GSM8K: Gemma-4-27B-A4B confirms the linear relationship ($\alpha{=}7.36$, $r{=}0.773$, $p{=}0.009$); Phi and Qwen deviate for the architecture-specific reasons discussed in Section~\ref{sec:routing_patterns} and Section~\ref{sec:qwen_dive} respectively, consistent with the Huffman signal being unmeasurable in those regimes.
Equation~\eqref{eq:temporal_decay} thus characterizes efficient MoE routing as a temporal compression process that continuously adapts code length to the evolving information content of the reasoning trajectory.

\paragraph{Architecture-Specific Routing Patterns.}
\label{sec:routing_patterns}
While the Frequency-Diversity Law serves as a universal governing principle, its empirical manifestation is strictly modulated by the expert sparsity ratio $k/E$ (Table~\ref{tab:routing_summary}). 

\textbf{Gemma-4-27B-A4B} ($k/E{=}6.25\%$) exhibits the most direct form of temporal compression, characterized by a clean monotonic decline in the unique expert count $u_i$; in this regime, early and late reasoning steps are distinguishable from the raw signal. 

\textbf{Phi-3.5-MoE} ($k/E{=}12.5\%$) presents a more complex temporal trajectory due to its inherent expert-index clustering (Table~\ref{tab:routing_summary}); here, temporal compression manifests as the routing state settling into stable, low-entropy cluster pairs rather than a simple reduction in cardinality. 

In contrast, \textbf{Qwen3.5-35B-A3B} employs the sparsest activation ratio ($k/E{=}3.1\%$), where the Huffman signal becomes statistically unmeasurable. As Section~\ref{sec:qwen_dive} reveals, the root cause is \emph{functional redundancy}: the vast majority of Qwen's 256 experts are near-duplicates, so routing codes expend bits selecting \emph{which equivalent expert} rather than encoding \emph{which semantic operation}. Table~\ref{tab:routing_summary} summarizes the three routing regimes.

\paragraph{Quantitative Validation.}
Figure~\ref{fig:huffman_scatter} reports the relationship between operation-type frequency $p(o)$ and mean unique expert count $\bar{u}^{(o)}$ (Eq.~\ref{eq:huffman}) across all three architectures.

For Gemma-4-27B-A4B and Phi-3.5-MoE, the four operation types conform to the predicted Huffman correspondence: both models attain Spearman $\rho = 1.00$ ($p < 0.05$). As shown in Figure~\ref{fig:huffman_scatter}, the mean expert count $\bar{u}^{(o)}$ for each operation type coincides with the reverse frequency ranking ($-\log p(o)$), consistent with Eq.~\eqref{eq:huffman}. Qwen3.5-35B-A3B does not follow this pattern; we defer its analysis to Section~\ref{sec:qwen_dive}, where its extreme activation sparsity ($k/E = 3.1\%$) is shown to induce a qualitatively distinct routing regime.
A further corroborating finding is reported in the appendix: routing fingerprints are content-independent (intra-type Jaccard $>$ inter-type across all 12 model--dataset pairs, $p \leq 10^{-163}$; Appendix~\ref{app:specialization}).

\section{Diving Deeper into Qwen: Expert Redundancy and Effective sparsity}
\label{sec:qwen_dive}

The analysis in Section~\ref{sec:routing_patterns} attributes Qwen3.5-35B-A3B's unmeasurable Huffman signal to its extreme routing sparsity.
This raises a natural follow-up question: \textit{although 256 experts \textbf{formally} exist, are all of them functionally distinct?}
If many experts are redundant (activating on the same operation types) then the \emph{effective} expert pool may be much smaller than 256, and the true $k/E$ ratio correspondingly higher.

\subsection{Discovering Expert Redundancy via Co-Activation Profiles}
\label{sec:qwen_redundancy}

To test this hypothesis, we build a \textbf{14-dimensional co-activation profile} for every (layer, expert) pair from correct CoT traces across four datasets (GSM8K \citep{cobbe2021training}, MATH \citep{hendrycks2021measuring}, AQuA \citep{ling2017program}, CompMath-MCQ \citep{lo2025closer}).
Each profile encodes 4 operation-type dimensions (fraction of activations per semantic type: \texttt{problem setup}, \texttt{add/subtract}, \texttt{multiply/divide}, \texttt{other}) and 10 stage dimensions (fraction per decile of the normalized trajectory, step $i/(n{-}1)$), normalized independently so both aspects contribute equally (construction details in Appendix~\ref{app:qwen_sim}).

We then compute the pairwise cosine similarity between all expert profiles within each layer.
Table~\ref{tab:qwen_sim} in Appendix~\ref{app:qwen_sim} reports the resulting similarity statistics across all 40 layers.

The figures are striking: across every layer, the \emph{average} pairwise cosine similarity between expert profiles exceeds 0.886, and the 90th-percentile similarity exceeds 0.983.
In other words, the vast majority of Qwen's 256 experts are nearly indistinguishable in terms of \emph{which operation types they activate on} and \emph{when in the reasoning chain they fire}.
Despite load-balancing loss during training forcing all experts to be approximately equally utilized, their functional roles (as captured by temporal co-activation structure) are highly redundant.

\subsection{Pruning Redundant Experts and Testing for Huffman Emergence}
\label{sec:qwen_pruning}

\paragraph{Pruning method.}
We identify redundant expert pairs via a greedy similarity-based procedure applied independently per layer:
\begin{enumerate}[leftmargin=1.5em,itemsep=0pt,topsep=2pt]
\item Compute the $256{\times}256$ cosine similarity matrix from co-activation profiles.
\item Iteratively find the most similar active pair $(i,j)$; remove the expert with the lower total activation frequency.
\item Repeat until the target pruning fraction is reached.
\end{enumerate}
At inference time, the removed experts are masked via a forward hook on the gate module: their router logits are set to $-\infty$ before top-$k$ selection, so they are never dispatched to.
No weights are modified; the procedure is entirely training-free.
Full algorithmic details are given in Appendix~\ref{app:pruning_method}.
If the redundancy hypothesis is correct, pruning should incur minimal accuracy loss and crucially raise $k/E_\text{eff}$ enough to make the Huffman signal detectable (see Appendix~\ref{app:qwen_pruning} for effective sparsity at each level).
Table~\ref{tab:qwen_pruning_acc} reports accuracy and Huffman $r$ on GSM8K under each condition.

\begin{table}[b]
\caption{%
    GSM8K accuracy and Huffman correlation for Qwen3.5-35B-A3B under SDP.
    % ``10\% expert pruned'' removes experts ranked 26--51 by redundancy
    % ($\text{set}_{20\%}\setminus\text{set}_{10\%}$, $E_\text{eff}=230$);
    % ``20\% expert pruned'' removes experts ranked 26--76
    % ($\text{set}_{30\%}\setminus\text{set}_{10\%}$, $E_\text{eff}=205$).
    % Huffman $r$: Pearson correlation between $-\log p(o)$ and $\bar{u}^{(o)}$ across four operation types.
}
\label{tab:qwen_pruning_acc}
\centering\small
\begin{tabular}{lcccc}
\toprule
Pruning & $E_\text{eff}$ & GSM8K Acc. & Huffman $r$ & Avg.\ Tokens\\
\midrule
0\% (baseline)      & 256 & 93.6\% & \textcolor{red}{$-0.631$} & 1{,}658 \\
10\% expert pruned  & 230 & 91.9\% & \textcolor{ForestGreen}{$+0.567$} & 1{,}899 \\
20\% expert pruned  & 205 & 75.7\% & \textcolor{ForestGreen}{$+0.533$} & 2{,}339 \\
\bottomrule
\end{tabular}
\end{table}

\begin{figure}[t]
    \centering
    \small \textbf{Qwen3.5-35B-A3B --- Huffman test: $\bar{u}^{(o)} \propto -\log p(o)$} \\ [1.0ex]
    \includegraphics[width=0.8\linewidth]{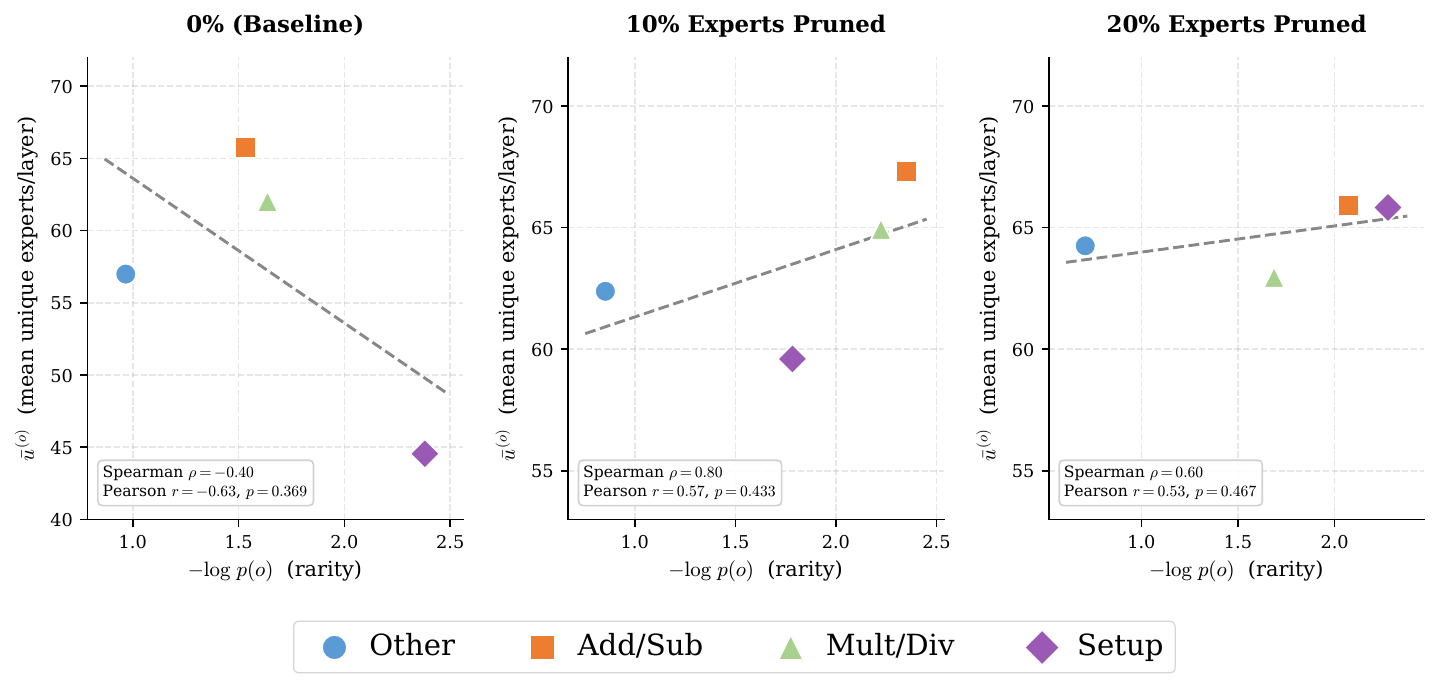}
    \caption{Huffman scatter plots for Qwen3.5-35B-A3B under 0\% (baseline), 10\%, and 20\%
    Subset Difference Pruning, evaluated on GSM8K.
    % Each point represents one of four operation types; the dashed line is an OLS fit.
    % The Pearson $r$ flips from $-0.631$ (anti-Huffman) to $+0.57/+0.53$ (Huffman-compliant)
    % after pruning mid-tier redundant experts, confirming that effective sparsity
    % governs whether the Huffman routing law is observable.
    }
    \label{fig:qwen_huffman}
\end{figure}

\paragraph{Results.}
Figure~\ref{fig:qwen_huffman} shows the Huffman scatter for all three conditions evaluated on GSM8K dataset.
The baseline (0\%, 256 experts) yields Pearson $r=-0.631$, indicating that \emph{frequent} operation types activate \emph{more} unique experts per step, which is an anti-Huffman signature.
The root cause is Qwen's \emph{inverted-U trajectory} (Figure~\ref{fig:unique_expert_comparison}): unlike Gemma and Phi, which show monotonic decline, Qwen's unique expert count rises during mid-trace exploration then collapses as the model converges toward a solution.
Because operation types are not uniformly distributed across trace positions; for instance, \texttt{problem setup} keywords (\emph{let, given  that, define, etc.}) appear disproportionately in the lower-$\bar{u}$ phases while arithmetic steps dominate the high-$\bar{u}$ exploration peak, \emph{temporal position within the trace}, rather than operation-type frequency, becomes the primary predictor of expert set size.
Consequently, the rarest operation type (\emph{setup}, $p{=}0.095$) receives the \emph{lowest} mean expert count ($\bar{u}{=}45.3$), directly inverting the Huffman expectation (full per-type statistics in Appendix~\ref{app:huffman_validation}).
After masking the middle-redundancy tier, both pruned conditions flip to positive correlation ($r\!=\!+0.57$ at 10\% pruning; $r\!=\!+0.53$ at 20\%), confirming that the Huffman phenomenon emerges once effective sparsity is raised.
The 10\%-pruned model retains near-baseline accuracy (91.9\% vs.\ 93.6\%), suggesting this level sits near the boundary of tolerable redundancy removal, whereas 20\% pruning incurs a more substantial drop (75.7\%), likely because the router can no longer adequately compensate for the missing capacity on harder problems.

\begin{figure}[t]
    \centering
     \small \textbf{Pruned Expert Proportion --- Qwen3.5-35B-A3B} \\ [1.0ex]
    \includegraphics[width=0.75\linewidth]{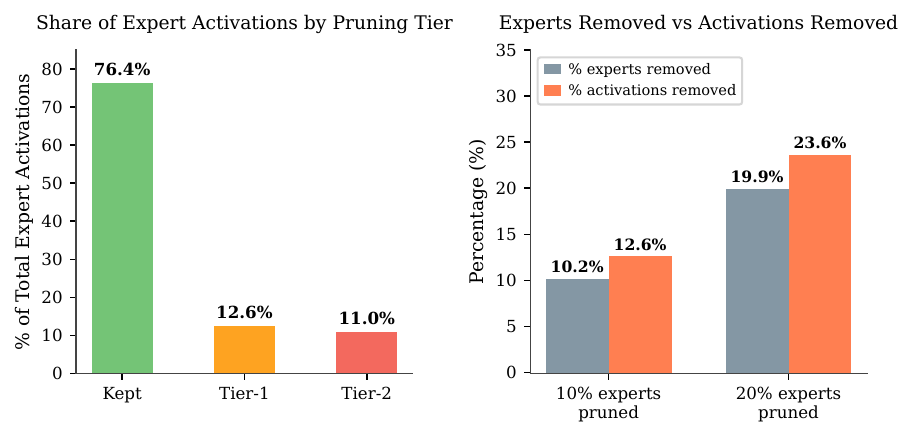}
    \caption{Left: Share of total expert activations partitioned by redundancy tier.
    Tier-1 (top 10\% most redundant experts, ${\sim}26$ per layer) accounts for 12.6\% of activations, while Tier-2 (the subsequent 10\% tier) contributes an additional 11.0\%.
    Right: Comparison of expert count removed versus activations lost for each pruning condition. The ``10\% expert pruned'' case removes Tier-1 (10.2\% of pool), whereas the ``20\% expert pruned'' case removes the combined Tier-1+2 (19.9\% of pool).
    In both settings, the share of lost activations exceeds the share of removed experts, confirming that these experts are actively utilized functional redundancies rather than inactive artifacts.}
    \label{fig:qwen_proportion}
\end{figure}

\subsection{What Gets Pruned, and Why It Matters}
\label{sec:qwen_pruned_analysis}

\paragraph{Activation share of pruned experts.}
The SDP algorithm iteratively removes the lower-frequency member from the most similar expert pairs. Figure~\ref{fig:qwen_proportion} compares activation loss across tiers: the Tier-1 experts (10.2\% of pool) account for \textbf{12.6\%} of activations, while the combined Tier-1+2 (19.9\% of pool) accounts for \textbf{23.6\%}. Because the activation share consistently exceeds the expert count share, these units are clearly not dormant relics; rather, they are active but functionally redundant components. At inference time, the router successfully redistributes their traffic to surviving experts within the same functional cluster, maintaining coverage despite the reduced architectural flexibility.

\begin{figure}[t]
    \centering
    \small \textbf{Expert Functional Profile t-SNE --- Qwen3.5-35B-A3B} \\ [1.0ex]
    \includegraphics[width=0.75\linewidth]{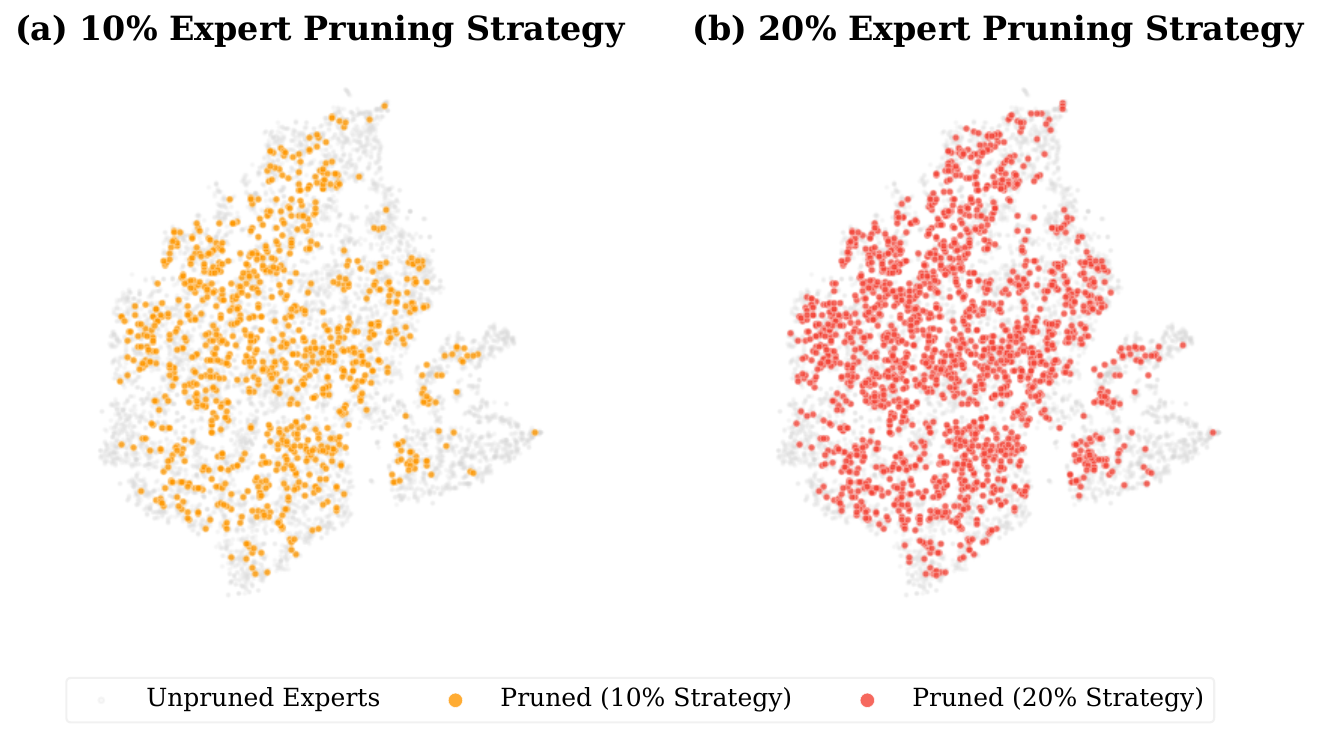}
    \caption{t-SNE visualization of expert functional profiles in Qwen3.5-35B-A3B. The 14-dimensional profiles are computed from 782 traces across all active (layer, expert) pairs. Panels illustrate experts removed via (left) 10\% Subset Difference Pruning (orange, 1,040 pairs) and (right) 20\% pruning (red, 2,040 pairs), overlaid on retained experts (grey). The uniform distribution of pruned experts suggests functional redundancy across the entire routing space.}
    \label{fig:qwen_tsne}
\end{figure}

\paragraph{Functional profile t-SNE.}
To verify that pruned experts do not occupy a narrow, easily-separable functional niche,
Figure~\ref{fig:qwen_tsne} plots t-SNE embeddings of the 14-dimensional functional profiles for all 10{,}185 active (layer, expert) pairs.
Tier-1 (orange) and tier-1+2 (red) points are \textbf{uniformly scattered} across the entire embedding and show no isolated cluster that separates them from the retained population.
This geometric evidence reinforces the redundancy interpretation: the removed experts are near-duplicate copies of retained experts, distributed throughout every functional region of the routing space rather than concentrated in any single operation-type or temporal niche.
Taken together, the activation-share and t-SNE results confirm that Qwen's 256-expert pool contains substantial built-in redundancy and that Subset Difference Pruning selectively removes it while preserving functional diversity.

From an information-theoretic perspective, this redundancy is a symptom of \emph{insufficient routing information density}: a large share of each routing codeword's bits are consumed identifying \emph{which functionally identical expert} was selected, rather than encoding \emph{which semantic operation} is being performed.
This explains both the anti-Huffman signature (Table~\ref{tab:qwen_pruning_acc}) and the diffuse t-SNE geometry: when many experts are interchangeable, the router's choices carry low mutual information with the operation type, and the resulting routing fingerprints are scattered rather than clustering semantically by operation.

\paragraph{Huffman Compliance as a Routing Health Diagnostic.}
The Huffman correlation $r$ (Eq.~\ref{eq:huffman}) provides a lightweight, task-free probe for expert pool quality: a positive $r$ confirms the router is operating as an MDL-optimal encoder, while a negative $r$ signals a \emph{redundancy trap}, where the effective sparsity $k/E_\text{eff}$ is too low for the Huffman law to manifest, typically because load-balancing has filled the expert pool with functional near-duplicates.
Crucially, this diagnostic requires only routing traces on a small held-out set and no accuracy benchmarks, making it applicable at any point in training or at deployment to assess whether the expert pool's functional diversity is properly aligned with the model's semantic operation distribution.

\section{Conclusion}

This study confirms that the MoE routing mechanism functions as a spontaneously evolved \textbf{Huffman Encoder}. Addressing the central hypothesis posed in the introduction: \emph{Routing is Compression}, our findings yield three key insights:

\begin{itemize}[leftmargin=1.5em,itemsep=0pt,topsep=2pt]
    \item \textbf{Universality of the Huffman Phenomenon:} In models like Gemma-4-27B-A4B and Phi-3.5-MoE, routing decisions exhibit peak information-theoretic efficiency. The monotonic increase of expert diversity with task rarity ($-\log p(o)$), evidenced by a Spearman's $\rho = 1.00$, proves that highly optimized models naturally allocate longer routing paths to complex semantic logic to minimize the MDL.
    
    \item \textbf{Antagonism between Redundancy and Efficiency:} The anomaly in Qwen3.5-35B-A3B reveals the pitfalls of aggressive load-balancing. Forced utilization diversity often results in functional redundancy (pairwise cosine similarity $> 0.90$), which obscures the latent Huffman signal and deviates from optimal source coding principles.
    
    \item \textbf{Structural Re-encoding via Pruning:} Through \textbf{Subset Difference Pruning}, we successfully removed the intermediate-redundancy tier of experts. This intervention reshaped Qwen's reasoning trajectories from an anti-Huffman state to an efficient encoding path (Pearson $r$ improved to $+0.57$) with negligible performance loss ($-1.7\%$ on GSM8K).
\end{itemize}

In conclusion, routing efficiency hinges on effective sparsity ($k/E_{\text{eff}}$). By maintaining a 5--13\% ratio, MoE routers transcend mere load-balancing to act as genuine semantic compression engines. This validates the MDL principle as a governing law for routing paths, \textbf{offering a theoretical blueprint for building high-performance, low-redundancy reasoning models.}
Beyond post-hoc analysis, the Huffman correlation $r$ opens a path toward training-time self-monitoring: if $r$ turns persistently negative during training, it signals that load balancing is generating functional duplicates faster than the model develops meaningful expert specialization---a diagnostic that requires no downstream evaluation.
Limitations are discussed in Appendix~\ref{app:limitations}.

\newpage
\bibliography{example_paper}

@inproceedings{lepikhin2021gshard,
  title     = {{GShard}: Scaling Giant Models with Conditional Computation and Automatic Sharding},
  author    = {Lepikhin, Dmitry and Lee, HyoukJoong and Xu, Yuanzhong and Chen, Dehao and Firat, Orhan and Huang, Yanping and Krikun, Maxim and Shazeer, Noam and Chen, Zhifeng},
  booktitle = {International Conference on Learning Representations (ICLR)},
  year      = {2021}
}

@article{jiang2024mixtral,
  title   = {Mixtral of Experts},
  author  = {Jiang, Albert Q and Sablayrolles, Alexandre and Roux, Antoine and Mensch, Arthur and Savary, Blanche and Bamford, Chris and Chaplot, Devendra Singh and de las Casas, Diego and Hanna, Emma Bou and others},
  journal = {arXiv preprint arXiv:2401.04088},
  year    = {2024}
}

@article{abdin2024phi,
  title   = {Phi-3 Technical Report: A Highly Capable Language Model Locally on Your Phone},
  author  = {Abdin, Marah and Aneja, Jyoti and Awadalla, Hany and Awadallah, Ahmed and Awan, Ammar Ahmad and Bach, Nguyen and Bahree, Amit and Bakhtiari, Arash and Bao, Jianmin and others},
  journal = {arXiv preprint arXiv:2404.14219},
  year    = {2024}
}

@article{yang2025qwen3,
  title={Qwen3 technical report},
  author={Yang, An and Li, Anfeng and Yang, Baosong and Zhang, Beichen and Hui, Binyuan and Zheng, Bo and Yu, Bowen and Gao, Chang and Huang, Chengen and Lv, Chenxu and others},
  journal={arXiv preprint arXiv:2505.09388},
  year={2025}
}

@article{Kamath2025Gemma3T,
  title={Gemma 3 Technical Report},
  author={Gemma Team Aishwarya Kamath and Johan Ferret and Shreya Pathak and Nino Vieillard and Ramona Merhej and Sarah Perrin and Tatiana Matejovicova and Alexandre Ram'e and Morgane Rivi{\`e}re and Louis Rouillard and Thomas Mesnard and Geoffrey Cideron and Jean-Bastien Grill and Sabela Ramos and Edouard Yvinec and Michelle Casbon and Etienne Pot and Ivo Penchev and Gael Liu and Francesco Visin and Kathleen Kenealy and Lucas Beyer and Xiaohai Zhai and Anton Tsitsulin and R{\'o}bert Istvan Busa-Fekete and Alex Feng and Noveen Sachdeva and Benjamin Coleman and Yi Gao and Basil Mustafa and Iain Barr and Emilio Parisotto and David Tian and Matan Eyal and Colin Cherry and Jan-Thorsten Peter and Danila Sinopalnikov and Surya Bhupatiraju and Rishabh Agarwal and Mehran Kazemi and Dan Malkin and Ravin Kumar and David Vilar and Idan Brusilovsky and Jiaming Luo and Andreas Steiner and Abe Friesen and Abhanshu Sharma and Abheesht Sharma and Adi Mayrav Gilady and Adrian Goedeckemeyer and Alaa Saade and Alexander Kolesnikov and Alexei Bendebury and Alvin Abdagic and Amit Vadi and Andr'as Gyorgy and Andr{\'e} Susano Pinto and Anil Das and Ankur Bapna and Antoine Miech and Antoine Yang and Antonia Paterson and Ashish Shenoy and Ayan Chakrabarti and Bilal Piot and Boxi Wu and Bobak Shahriari and Bryce Petrini and Charlie Chen and Charline Le Lan and Christopher A. Choquette-Choo and Cj Carey and Cormac Brick and Daniel Deutsch and Danielle Eisenbud and Dee Cattle and Derek Cheng and Dimitris Paparas and Divyashree Shivakumar Sreepathihalli and Doug Reid and Dustin Tran and Dustin Zelle and Eric Noland and Erwin Huizenga and Eugene Kharitonov and Frederick Liu and Gagik Amirkhanyan and Glenn Cameron and Hadi Hashemi and Hanna Klimczak-Pluci'nska and Harman Singh and Harsh Mehta and Harshal Tushar Lehri and Hussein Hazimeh and Ian Ballantyne and Idan Szpektor and Ivan Nardini and Jean Pouget-Abadie and Jetha Chan and Joe Stanton and J. Michael Wieting and Jonathan Lai and Jordi Orbay and Joe Fernandez and Joshua Newlan and Junsong Ji and Jyotinder Singh and Kat Black and Kathy Yu and Kevin Hui and Kiran Vodrahalli and Klaus Greff and Linhai Qiu and Marcella Valentine and Marina Coelho and Marvin Ritter and Matt Hoffman and Matthew Watson and Mayank Chaturvedi and Michael Moynihan and Min Ma and Nabila Babar and Natasha Noy and Nathan Byrd and Nick Roy and Nikola Momchev and Nilay Chauhan and Oskar Bunyan and Pankil Botarda and Paul Caron and Paul Kishan Rubenstein and Phil Culliton and Philipp Schmid and Pier Giuseppe Sessa and Ping-mei Xu and Piotr Stańczyk and Pouya Dehghani Tafti and Rakesh Shivanna and Renjie Wu and Renke Pan and Reza Ardeshir Rokni and Rob Willoughby and Rohith Vallu and Ryan Mullins and Sammy Jerome and Sara Smoot and Sertan Girgin and Shariq Iqbal and Shashir Reddy and Shruti Sheth and Siim P{\~o}der and Sijal Bhatnagar and Sindhu Raghuram Panyam and Sivan Eiger and Susan Zhang and Tianqi Liu and Trevor Yacovone and Tyler Liechty and Uday Kalra and Utku Evci and Vedant Misra and Vincent Roseberry and Vladimir Feinberg and Vlad Kolesnikov and Woohyun Han and Woosuk Kwon and Xi Chen and Yinlam Chow and Yuvein Zhu and Zichuan Wei and Zoltan Egyed and Victor Cotruta and Minh Giang and Phoebe Kirk and Anand Rao and Jessica Lo and Erica Moreira and Luiz Gustavo Martins and Omar Sanseviero and Lucas Gonzalez and Zach Gleicher and Tris Warkentin and Vahab S. Mirrokni and Evan Senter and Eli Collins and Joelle Barral and Zoubin Ghahramani and Raia Hadsell and Yossi Matias and D. Sculley and Slav Petrov and Noah Fiedel and Noam Shazeer and Oriol Vinyals and Jeffrey Dean and Demis Hassabis and Koray Kavukcuoglu and Cl{\'e}ment Farabet and Elena Buchatskaya and Jean-Baptiste Alayrac and Rohan Anil and Dmitry Lepikhin and Sebastian Borgeaud and Olivier Bachem and Armand Joulin and Alek Andreev and Cassidy Hardin and Robert Dadashi and L'eonard Hussenot},
  journal={ArXiv},
  year={2025},
  volume={abs/2503.19786},
  url={https://api.semanticscholar.org/CorpusID:277313563}
}

@article{cobbe2021training,
  title   = {Training Verifiers to Solve Math Word Problems},
  author  = {Cobbe, Karl and Kosaraju, Vineet and Bavarian, Mohammad and Chen, Mark and Jun, Heewoo and Kaiser, Lukasz and Plappert, Matthias and Tworek, Jerry and Hilton, Jacob and Nakano, Reiichiro and others},
  journal = {arXiv preprint arXiv:2110.14168},
  year    = {2021}
}

@inproceedings{hendrycks2021measuring,
  title     = {Measuring Mathematical Problem Solving with the {MATH} Dataset},
  author    = {Hendrycks, Dan and Burns, Collin and Kadavath, Saurav and Arora, Akul and Basart, Steven and Tang, Eric and Song, Dawn and Steinhardt, Jacob},
  booktitle = {Advances in Neural Information Processing Systems (NeurIPS)},
  year      = {2021}
}

@inproceedings{ling2017program,
  title     = {Program Induction by Rationale Generation: Learning to Solve and Explain Algebraic Word Problems},
  author    = {Ling, Wang and Yogatama, Dani and Dyer, Chris and Blunsom, Phil},
  booktitle = {Proceedings of the Association for Computational Linguistics (ACL)},
  year      = {2017}
}

@INPROCEEDINGS{zoph2022designing,
  author={Zoph, Barret},
  booktitle={2022 IEEE International Parallel and Distributed Processing Symposium Workshops (IPDPSW)}, 
  title={Designing Effective Sparse Expert Models}, 
  year={2022},
  volume={},
  number={},
  pages={1044-1044},
  doi={10.1109/IPDPSW55747.2022.00171}}

@inproceedings{artetxe2022efficient,
  title     = {Efficient Large Scale Language Modeling with Mixtures of Experts},
  author    = {Artetxe, Mikel and Bhosale, Shruti and Goyal, Naman and Mihaylov, Todor and Ott, Myle and Shleifer, Sam and Lin, Xi Victoria and Du, Jingfei and Iyer, Srinivasan and Pasunuru, Ramakanth and others},
  booktitle = {Proceedings of the Conference on Empirical Methods in Natural Language Processing (EMNLP)},
  year      = {2022}
}

@inproceedings{geva2021transformer,
  title     = {Transformer Feed-Forward Layers Are Key-Value Memories},
  author    = {Geva, Mor and Schuster, Roei and Berant, Jonathan and Levy, Omer},
  booktitle = {Proceedings of the Conference on Empirical Methods in Natural Language Processing (EMNLP)},
  year      = {2021}
}

@article{elhage2022solu,
  title   = {Softmax Linear Units},
  author  = {Elhage, Nelson and Hume, Tristan and Olsson, Catherine and Nanda, Neel and Henighan, Tom and Johnston, Scott and ElShowk, Sheer and Joseph, Nicholas and DasSarma, Nova and Mann, Ben and others},
  journal = {Transformer Circuits Thread},
  year    = {2022},
  note    = {\url{https://transformer-circuits.pub/2022/solu/index.html}}
}

@article{zhou2025landscape,
  title   = {Landscape of Thoughts: Visualizing the Reasoning Process of Large Language Models},
  author  = {Zhou, Zhanke and Zhu, Zhuoqing and Li, Xiao and Galkin, Mikhail and Feng, Xiao and Koyejo, Sanmi and Han, Bo},
  journal = {arXiv preprint arXiv:2503.22165},
  year    = {2025}
}

@inproceedings{han2015learning,
  title     = {Learning both Weights and Connections for Efficient Neural Networks},
  author    = {Han, Song and Pool, Jeff and Tran, John and Dally, William J.},
  booktitle = {Advances in Neural Information Processing Systems},
  volume    = {28},
  year      = {2015}
}

@article{frantar2022gptq,
  title   = {{GPTQ}: Accurate Post-Training Quantization for Generative
             Pre-trained Transformers},
  author  = {Frantar, Elias and Ashkboos, Saleh and Hoefler, Torsten and
             Alistarh, Dan},
  journal = {arXiv preprint arXiv:2210.17323},
  year    = {2022}
}

@article{rissanen1978modeling,
  title   = {Modeling by Shortest Data Description},
  author  = {Rissanen, Jorma},
  journal = {Automatica},
  volume  = {14},
  number  = {5},
  pages   = {465--471},
  year    = {1978}
}

@inproceedings{deletang2023language,
  title     = {Language Modeling Is Compression},
  author    = {Del{\'e}tang, Gr{\'e}goire and Ruoss, Anian and Duquenne,
               Paul-Ambroise and Catt, Elliot and Genewein, Tim and
               Mattern, Christopher and Grau-Moya, Jordi and Wenliang, Li Kevin
               and Aitchison, Matthew and Orseau, Laurent and Hutter, Marcus
               and Veness, Joel},
  booktitle = {International Conference on Learning Representations},
  year      = {2024}
}

@book{cover2006elements,
  title     = {Elements of Information Theory},
  author    = {Cover, Thomas M. and Thomas, Joy A.},
  edition   = {2nd},
  year      = {2006},
  publisher = {Wiley-Interscience}
}

@article{krajewski2024scaling,
  title={Scaling laws for fine-grained mixture of experts},
  author={Krajewski, Jakub and Ludziejewski, Jan and Adamczewski, Kamil and Pi{\'o}ro, Maciej and Krutul, Micha{\l} and Antoniak, Szymon and Ciebiera, Kamil and Kr{\'o}l, Krystian and Odrzyg{\'o}{\'z}d{\'z}, Tomasz and Sankowski, Piotr and others},
  journal={arXiv preprint arXiv:2402.07871},
  year={2024}
}

@article{huffman1952method,
  title   = {A Method for the Construction of Minimum-Redundancy Codes},
  author  = {Huffman, David A.},
  journal = {Proceedings of the IRE},
  volume  = {40},
  number  = {9},
  pages   = {1098--1101},
  year    = {1952}
}

@article{Shannon1948method,
  title   = {A Mathematical Theory of Communication},
  author  = {Shannon, Claude E.},
  journal = {Bell System Technical Journal},
  volume  = {27},
  number  = {3},
  pages   = {379-423},
  year    = {1948}
}

@inproceedings{shazeer2017outrageously,
  title     = {Outrageously Large Neural Networks: The Sparsely-Gated Mixture-of-Experts Layer},
  author    = {Shazeer, Noam and Mirhoseini, Azalia and Maziarz, Krzysztof and Davis, Andy and Le, Quoc and Hinton, Geoffrey and Dean, Jeff},
  booktitle = {International Conference on Learning Representations (ICLR)},
  year      = {2017}
}

@book{hutter2005universal,
  title={Universal artificial intelligence: Sequential decisions based on algorithmic probability},
  author={Hutter, Marcus},
  volume={300},
  year={2005},
  publisher={Springer}
}

@article{shannon1959coding,
  title={Coding theorems for a discrete source with a fidelity criterion},
  author={Shannon, Claude E},
  journal = {IRE International Convention Record},
  pages={142-163},
  volumne={7},
  year={1959}
}

@article{fedus2022switch,
  title   = {Switch Transformers: Scaling to Trillion Parameter Models with Simple and Efficient Sparsity},
  author  = {Fedus, William and Zoph, Barret and Shazeer, Noam},
  journal = {Journal of Machine Learning Research},
  volume  = {23},
  number  = {120},
  pages   = {1--39},
  year    = {2022}
}

@inproceedings{dai2024deepseekmoe,
  title={Deepseekmoe: Towards ultimate expert specialization in mixture-of-experts language models},
  author={Dai, Damai and Deng, Chengqi and Zhao, Chenggang and Xu, RX and Gao, Huazuo and Chen, Deli and Li, Jiashi and Zeng, Wangding and Yu, Xingkai and others},
  booktitle={Proceedings of the 62nd Annual Meeting of the Association for Computational Linguistics (Volume 1: Long Papers)},
  pages={1280--1297},
  year={2024}
}

@inproceedings{clark2022unified,
  title={Unified scaling laws for routed language models},
  author={Clark, Aidan and de Las Casas, Diego and Guy, Aurelia and Mensch, Arthur and Paganini, Michela and Hoffmann, Jordan and Damoc, Bogdan and Hechtman, Blake and Cai, Trevor and Borgeaud, Sebastian and others},
  booktitle={International conference on machine learning},
  pages={4057--4086},
  year={2022},
  organization={PMLR}
}

@inproceedings{zhou2022mixture,
  title     = {Mixture-of-Experts with Expert Choice Routing},
  author    = {Zhou, Yanqi and Lei, Tao and Liu, Hanxiao and Du, Nan and Huang, Yanping and Zhao, Vincent and Dai, Andrew and Chen, Zhifeng and Le, Quoc and Laudon, James},
  booktitle = {Advances in Neural Information Processing Systems},
  volume    = {35},
  pages     = {7103--7114},
  year      = {2022},
  publisher = {Curran Associates, Inc.},
  url       = {https://proceedings.neurips.cc/paper_files/paper/2022/hash/2f00ecd787b432c1d36f3de9800728eb-Abstract-Conference.html}
}

@inproceedings{puigcerver2023sparse,
  title={From sparse to soft mixtures of experts},
  author={Puigcerver, Joan and Ruiz, Carlos Riquelme and Mustafa, Basil and Houlsby, Neil},
  booktitle={The Twelfth International Conference on Learning Representations},
  year={2023}
}

@article{herbst2026expert,
  title={The Expert Strikes Back: Interpreting Mixture-of-Experts Language Models at Expert Level},
  author={Herbst, Jeremy and Lee, Jae Hee and Wermter, Stefan},
  journal={arXiv preprint arXiv:2604.02178},
  year={2026}
}

@inproceedings{xue2024openmoe,
  title     = {{OpenMoE}: An Early Effort on Open Mixture-of-Experts Language Models},
  author    = {Xue, Fuzhao and Zheng, Zian and Fu, Yao and Ni, Jinjie and Zheng, Zangwei and Zhou, Wangchunshu and You, Yang},
  booktitle = {Proceedings of the 41st International Conference on Machine Learning},
  series    = {Proceedings of Machine Learning Research},
  volume    = {235},
  pages     = {55625--55655},
  year      = {2024},
  publisher = {PMLR}
}

@article{mi2026effective,
  title={Effective MoE-based LLM Compression by Exploiting Heterogeneous Inter-Group Experts Routing Frequency and Information Density},
  author={Mi, Zhendong and Chen, Yixiao and Zhao, Pu and Yu, Xiaodong and Wang, Hao and Wang, Yanzhi and Huang, Shaoyi},
  journal={arXiv preprint arXiv:2602.09316},
  year={2026}
}

@inproceedings{lo2025closer,
  title={A closer look into mixture-of-experts in large language models},
  author={Lo, Ka Man and Huang, Zeyu and Qiu, Zihan and Wang, Zili and Fu, Jie},
  booktitle={Findings of the Association for Computational Linguistics: NAACL 2025},
  pages={4427--4447},
  year={2025}
}

@article{zheng2026unveiling,
  title={Unveiling Language Routing Isolation in Multilingual MoE Models for Interpretable Subnetwork Adaptation},
  author={Zheng, Kening and Huang, Wei-Chieh and Huo, Jiahao and Li, Zhonghao and Zou, Henry Peng and Yan, Yibo and Zou, Xin and Li, Jungang and Li, Junzhuo and Zhang, Hanrong and others},
  journal={arXiv preprint arXiv:2604.03592},
  year={2026}
}

@article{chen2026understanding,
  title={Understanding Multilingualism in Mixture-of-Experts LLMs: Routing Mechanism, Expert Specialization, and Layerwise Steering},
  author={Chen, Yuxin and Cai, Zhengzhou and Ji, Xiangtian and Zhao, Weixiang and Zhang, An and Wang, Xiang and Chua, Tat-Seng},
  journal={arXiv preprint arXiv:2601.14050},
  year={2026}
}

@article{wei2022chain,
  title={Chain-of-Thought Prompting Elicits Reasoning in Large Language Models},
  author={Wei, Jason and Wang, Xuezhi and Schuurmans, Dale and Bosma, Maarten and Ichter, Brian and Xia, Fei and Chi, Ed H. and Le, Quoc V and Zhou, Denny},
  journal={Advances in Neural Information Processing Systems},
  volume={35},
  pages={24824--24837},
  year={2022}
}

@article{vandermaaten2008visualizing,
  title={Visualizing data using t-SNE},
  author={Van der Maaten, Laurens and Hinton, Geoffrey},
  journal={Journal of machine learning research},
  volume={9},
  number={11},
  year={2008}
}

@article{omi2025load,
  title={Load balancing mixture of experts with similarity preserving routers},
  author={Omi, Nabil and Sen, Siddhartha and Farhadi, Ali},
  journal={arXiv preprint arXiv:2506.14038},
  year={2025}
}
\bibliographystyle{icml2026}

%%%%%%%%%%%%%%%%%%%%%%%%%%%%%%%%%%%%%%%%%%%%%%%%%%%%%%%%%%%%%%%%%%%%%%%%%%%%%%%
%%%%%%%%%%%%%%%%%%%%%%%%%%%%%%%%%%%%%%%%%%%%%%%%%%%%%%%%%%%%%%%%%%%%%%%%%%%%%%%
% APPENDIX
%%%%%%%%%%%%%%%%%%%%%%%%%%%%%%%%%%%%%%%%%%%%%%%%%%%%%%%%%%%%%%%%%%%%%%%%%%%%%%%
%%%%%%%%%%%%%%%%%%%%%%%%%%%%%%%%%%%%%%%%%%%%%%%%%%%%%%%%%%%%%%%%%%%%%%%%%%%%%%%
\newpage
\appendix
\onecolumn

\section{Model Architecture Details}
\label{app:arch}

Table~\ref{tab:arch} summarizes the key routing parameters for each model.

\begin{table}[H]
\caption{MoE architecture parameters used in this work.}
\label{tab:arch}
\centering
\begin{tabular}{lcccc}
\toprule
Model & \# Layers & \# Experts & Top-$k$ & Gate type \\
\midrule
Gemma-4-27B-A4B & 30 & 128 & 8 & Linear projection (softmax) \\
Phi-3.5-MoE & 32 & 16 & 2 & Softmax router \\
Qwen3.5-35B-A3B & 40 & 256 & 8 & Softmax gate \\
\bottomrule
\end{tabular}
\end{table}

%%%%%%%%%%%%%%%%%%%%%%%%%%%%%%%%%%%%%%%%%%%%%%%%%%%%%%%%%%%%

\section{Architecture-Specific Routing Patterns: Extended Discussion}
\label{app:routing_patterns}

The three models span qualitatively distinct routing regimes determined by their expert sparsity ratio $k/E$.

\textbf{Gemma-4-27B-A4B} activates $k{=}8$ out of $E{=}128$ experts ($k/E{=}6.25\%$), an intermediate sparsity that creates a natural balance between exploration and convergence.
Early reasoning steps engage a genuinely diverse expert set (high $u_s$), while late steps settle onto a smaller, stable subset; crucially, both behaviors are directly and monotonically reflected in the raw unique count $u_s$ without requiring normalization or smoothing.
The unique expert count and the temporal compression signal are, in Gemma-4's case, essentially the same measure, making it the clearest empirical demonstration of the Huffman temporal compression phenomenon.

\textbf{Phi-3.5-MoE} routes each token to exactly $k{=}2$ experts from $E{=}16$, giving $k/E{=}12.5\%$---the coarsest ratio among our models.
With only 16 total experts, individual tokens can cover a large fraction of the pool within a single step.
We observe a characteristic \emph{odd/even clustering} phenomenon: the router tends to pair experts from distinct index groups (e.g., $\{0,1\}$, $\{2,3\}$, \ldots) and alternates between these clusters across tokens.
This produces an \emph{oscillating} unique-expert trajectory rather than a monotonic decline.
Temporal compression in Phi-3.5-MoE therefore manifests as the model settling into a stable, repeating cluster pair in later steps rather than a gradual count reduction; detecting convergence requires observing the trajectory over a window of multiple steps rather than any single-step count.

\textbf{Qwen3.5-35B-A3B} ($k/E{=}3.1\%$) represents the sparsest regime. While its routing does show statistically significant operation-type specialization (Table~\ref{tab:specialization}), the Huffman frequency--diversity signal remains unmeasurable: temporal position within the trace dominates operation-type frequency as a predictor of $u_i$. As Section~\ref{sec:qwen_dive} reveals, the underlying cause is functional redundancy---near-duplicate experts are repeatedly co-activated regardless of semantic operation, so routing codes expend bits selecting \emph{which equivalent expert} rather than encoding \emph{which semantic operation}.

%%%%%%%%%%%%%%%%%%%%%%%%%%%%%%%%%%%%%%%%%%%%%%%%%%%%%%%%%%%%

\section{t-SNE Visualizations of Routing State}
\label{app:tsne}

\begin{figure}[H]
\centering
\small \textbf{t-SNE of Expert Routing Fingerprints} \\ [1.0ex]
\begin{subfigure}[b]{0.31\textwidth}
    \centering
    \includegraphics[width=\textwidth]{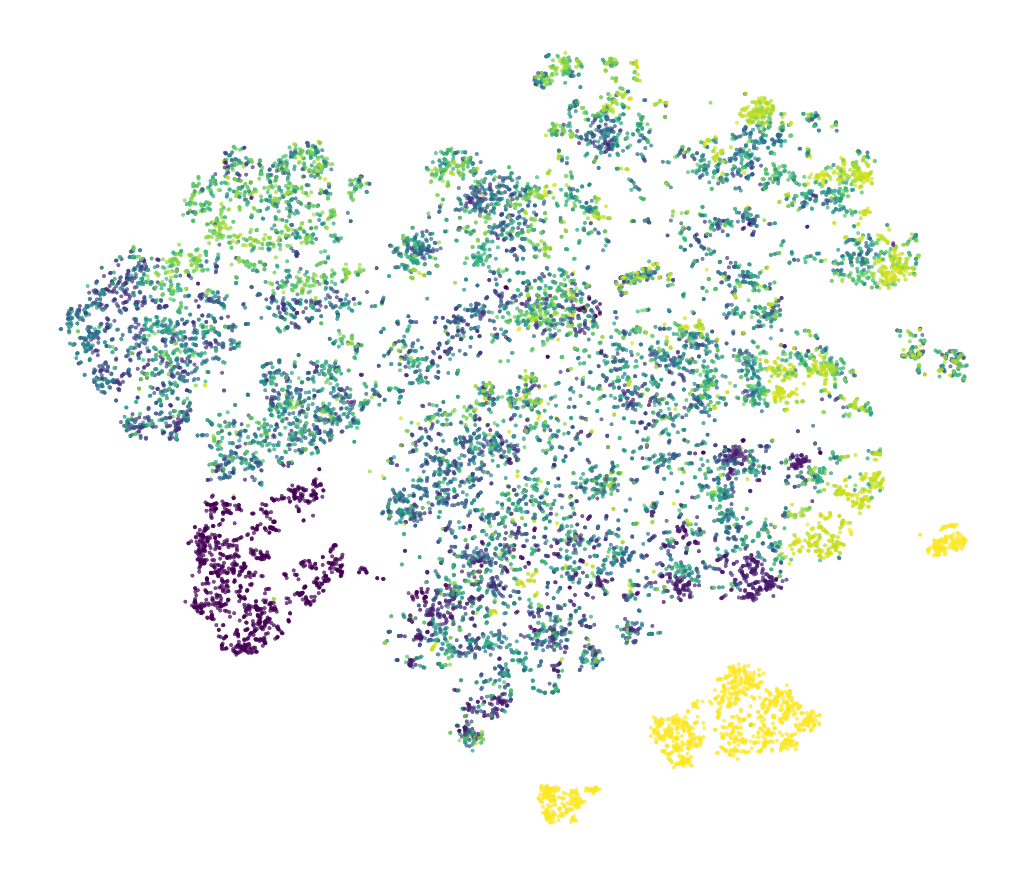}
    \caption{Gemma-4-27B-A4B}
    \label{fig:tsne:gemma}
\end{subfigure}
\hfill
\begin{subfigure}[b]{0.31\textwidth}
    \centering
    \includegraphics[width=\textwidth]{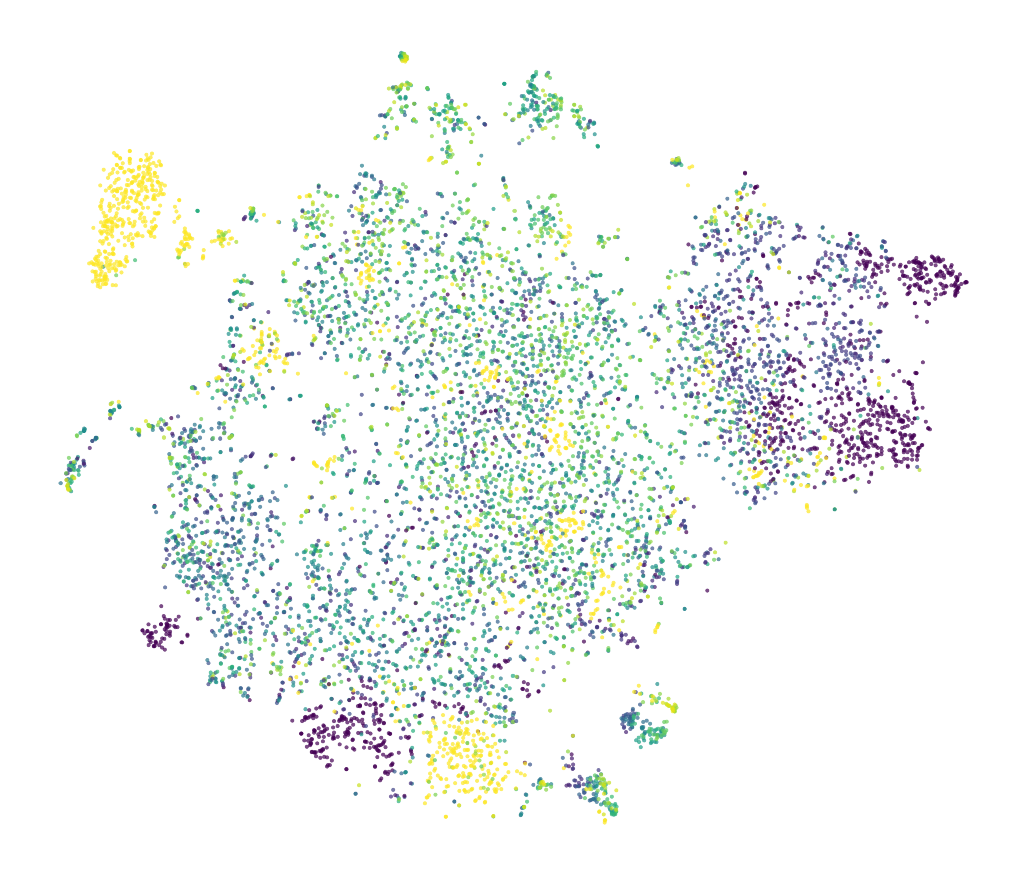}
    \caption{Phi-3.5-MoE}
    \label{fig:tsne:phi}
\end{subfigure}
\hfill
\begin{subfigure}[b]{0.31\textwidth}
    \centering
    \includegraphics[width=\textwidth]{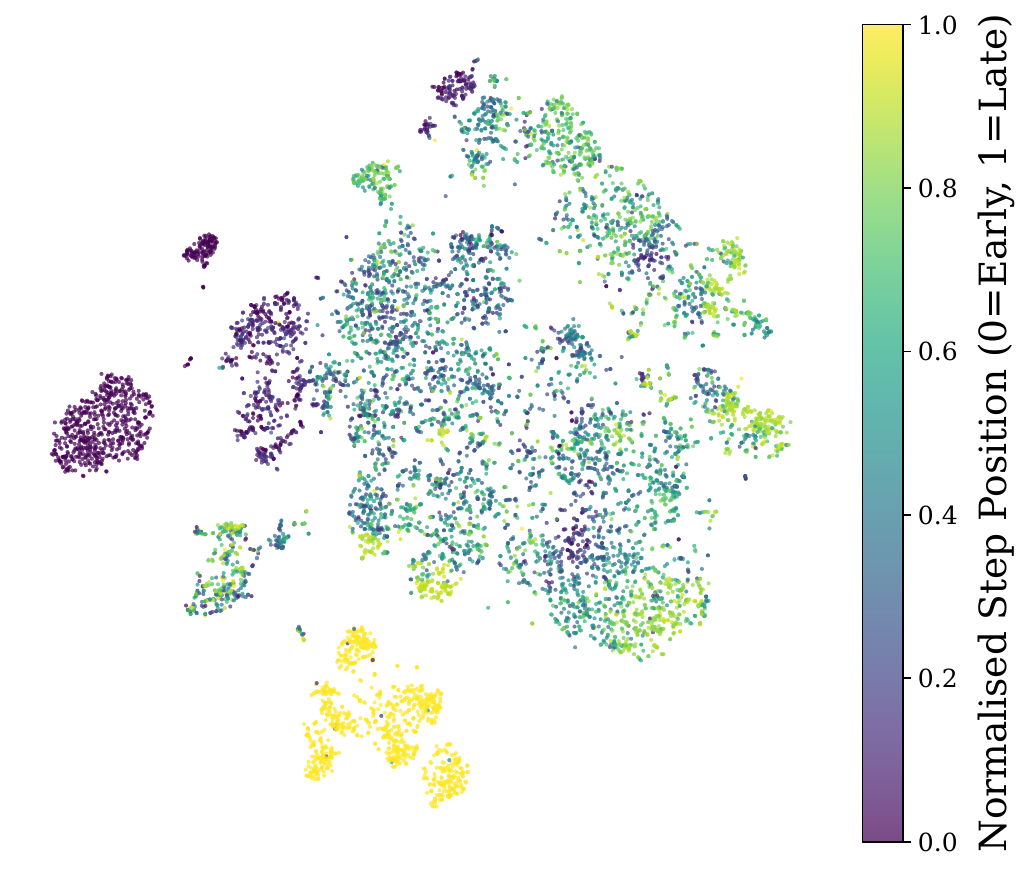}
    \caption{Qwen3.5-35B-A3B}
    \label{fig:tsne:qwen}
\end{subfigure}
\vspace{1ex}
\caption{t-SNE projections of expert routing fingerprints colored by normalized step position (0=early purple, 1=late yellow). While Gemma-4-27B-A4B and Phi-3.5-MoE exhibit smooth temporal transitions, Qwen3.5-35B-A3B displays distinct, isolated clusters at the extremes. These represent the \textbf{setup phase} (early purple cluster) and the \textbf{final answer output} (late yellow cluster), where the model utilizes the \textbf{minimum number of experts}. This high sparsity at the start and end of the reasoning trace results in highly specific routing fingerprints that manifest as extreme clusters, whereas the intermediate steps show scattered patterns due to functional redundancy among its experts.}
\label{fig:tsne_all}
\end{figure}

Figure~\ref{fig:tsne_all} shows t-SNE projections of the expert activation patterns for all three models. Each point represents the routing fingerprint of a single reasoning step, colored by its normalized position along the reasoning trace (a continuous gradient from early to late steps). The smooth color transitions visible in Gemma-4-27B-A4B and Phi-3.5-MoE provide geometric evidence that routing states evolve in an orderly temporal trajectory throughout the reasoning trace.

\paragraph{Feature construction and embedding procedure.}
For each reasoning step $i$, we construct a binary indicator vector $\mathbf{v}_i \in \{0,1\}^{L \times E}$ recording which experts were activated at each of the $L$ sensitive layers, where $E$ is the total number of experts per layer. To reduce dimensionality before embedding, we first apply PCA to retain 50 principal components, then apply t-SNE \citep{vandermaaten2008visualizing} with perplexity 30, 1{,}000 iterations, and random seed 42 (scikit-learn defaults otherwise). Points are colored by normalized step position along the reasoning trace, using a continuous colormap from early (position 0) to late (position 1) steps.

\paragraph{Geometric Corroboration of the Huffman Hypothesis.}
The cluster topology in Figure~\ref{fig:tsne_all} provides geometric support for the Frequency-Diversity Law.
In Gemma-4-27B-A4B and Phi-3.5-MoE, colors transition smoothly across the embedding from early to late positions, confirming that late-stage routing states occupy a qualitatively distinct, more concentrated region of the expert combination space---consistent with the convergence toward shorter combinatorial codes predicted by Huffman optimality.
The smoothness of this color gradient varies systematically across architectures in a manner directly attributable to their sparsity ratio $k/E$: higher sparsity (Phi, $k/E{=}12.5\%$) produces a sharper, more globally ordered gradient, while intermediate sparsity (Gemma, $k/E{=}6.25\%$) yields a more gradual but still coherent transition.
Qwen3.5-35B-A3B, with the sparsest ratio ($k/E{=}3.1\%$), shows the least spatial separation.
This is best understood through two compounding effects revealed by Section~\ref{sec:qwen_dive}.
First, the functional profiles of the vast majority of Qwen's 256 experts are nearly identical (pairwise cosine similarity $>{0.90}$ across all 40 layers): many experts specialize in precisely the same operation types and temporal stages, forming large redundant clusters.
Second, within each such cluster, the router has no principled reason to select the same subset of experts for the same operation type—it can route through any of several functionally equivalent members.
As a result, steps performing the same operation (e.g., multiplication) may activate entirely different expert subsets across different problems, even though those subsets are functionally interchangeable.
The routing fingerprints for same-operation steps are therefore scattered rather than clustered in t-SNE space—not because the temporal compression phenomenon is absent, but because the routing code is spending most of its bits encoding \emph{which of many equivalent experts} was chosen rather than \emph{which semantic operation} is being performed.
This is precisely the low information density diagnosis: the code length is high but the information content is not, because the expert pool contains far more functional redundancy than semantic diversity.
Removing the redundant middle tier via Subset Difference Pruning reduces the pool to more functionally distinct experts, concentrating information content per routing bit and allowing the Huffman signal to emerge.

%%%%%%%%%%%%%%%%%%%%%%%%%%%%%%%%%%%%%%%%%%%%%%%%%%%%%%%%%%%%

\section{Qwen3.5-35B-A3B Expert Functional Profile Analysis}
\label{app:qwen_sim}

\subsection*{Pairwise Cosine Similarity}

\begin{table}[ht]
\caption{%
    Pairwise cosine similarity between co-activation profiles of all 256 experts, aggregated across 40 layers of Qwen3.5-35B-A3B.
    All statistics are computed over the 782-trace corpus; ``active experts'' counts experts seen at least once.
}
\label{tab:qwen_sim}
\centering
\small
\begin{tabular}{lccccc}
\toprule
Statistic & Min & Mean & Max \\
\midrule
Active experts per layer & 243 & 254.6 & 256 \\
Pairwise sim.\ (mean)   & 0.886 & 0.908 & 0.949 \\
Pairwise sim.\ (P90)    & 0.983 & 0.987 & 0.993 \\
Pairwise sim.\ (P99)    & 0.994 & 0.996 & 0.998 \\
\bottomrule
\end{tabular}
\end{table}

\subsection*{t-SNE of Expert Functional Profiles by Pruning Tier}

To verify that pruned experts do not occupy a narrow, easily-separable functional niche, Figure~\ref{fig:qwen_tsne} (main text) plots t-SNE embeddings of the 14-dimensional functional profiles for all 10{,}185 active (layer, expert) pairs.
Tier-1 (orange) and tier-1+2 (red) points are \textbf{uniformly scattered} across the entire embedding and show no isolated cluster separating them from the retained population.
This geometric evidence reinforces the redundancy interpretation from the similarity table above: the removed experts are near-duplicate copies of retained experts, distributed throughout every functional region of the routing space rather than concentrated in any single operation-type or temporal niche.

Crucially, the scattering is not a simple consequence of high-dimensional sparsity.
The same functional profiles that yield pairwise cosine similarity $>{0.90}$ (Table~\ref{tab:qwen_sim}) also produce overlapping t-SNE clusters: because many experts specialize in precisely the same operation types and temporal stages, the router has no principled preference for any particular subset when routing a given step.
The routing code therefore spends bits encoding \emph{which of many equivalent experts} was selected rather than \emph{which semantic operation} is being performed---a direct symptom of insufficient routing information density.
Subset Difference Pruning reduces the pool to a more functionally distinct set, allowing each routing bit to carry more semantic information and enabling the Huffman signal to emerge.

%%%%%%%%%%%%%%%%%%%%%%%%%%%%%%%%%%%%%%%%%%%%%%%%%%%%%%%%%%%%
\section{Expert pruning levels for Qwen3.5-35B-A3B}
\label{app:qwen_pruning}
\begin{table}[H]
\caption{%
    Expert pruning levels for Qwen3.5-35B-A3B and the resulting effective sparsity ratios.
    The column $k/E_\text{eff}$ measures routing sparsity in the pruned model; Gemma-4's ratio (6.25\%) is provided for reference.
}
\label{tab:qwen_pruning}
\centering
\small
\begin{tabular}{ccccc}
\toprule
Pruning & Pruned / layer & $E_\text{eff}$ & $k/E_\text{eff}$ & Gemma-4 ref.\ \\
\midrule
10\% & 25  & 231 & 3.5\% & \multirow{4}{*}{6.25\%} \\
20\% & 51  & 205 & 3.9\% & \\
30\% & 76  & 180 & 4.4\% & \\
50\% & 128 & 128 & 6.2\% & \\
\bottomrule
\end{tabular}
\end{table}

%%%%%%%%%%%%%%%%%%%%%%%%%%%%%%%%%%%%%%%%%%%%%%%%%%%%%%%%%%%%

\section{Subset Difference Pruning: Algorithmic Details}
\label{app:pruning_method}

We identify redundant experts through a hierarchical similarity-based procedure applied independently per layer to preserve core model capabilities.

\begin{enumerate}[leftmargin=1.5em,itemsep=0pt,topsep=2pt]
\item \textbf{Profile similarity matrix.}
      Compute the $256{\times}256$ cosine similarity matrix from co-activation profiles (Section~\ref{sec:qwen_redundancy}) to quantify semantic redundancy between all expert pairs.
\item \textbf{Establish a Preservation Tier.}
      Designate the top 10\% of experts by total activation frequency (${\sim}25$ per layer) as a ``protected core'' that is exempt from pruning.
      These are the experts most consistently activated across all operation types and reasoning stages.
\item \textbf{Greedy Redundancy Removal.}
      Among the remaining unprotected experts, iteratively identify the most similar pair $(i, j)$ and remove the expert with the lower total activation frequency.
      This step removes functional near-duplicates while retaining the more informative member of each similar pair.
\item \textbf{Repeat until target fraction reached.}
      Continue greedy removal until the desired total pruning fraction (e.g., 20\% or 30\%) is achieved.
\end{enumerate}

At inference time, the removed experts are masked via a forward hook on the gate module: their router logits are set to $-\infty$ before top-$k$ selection, ensuring they are never dispatched to.
This forces the router to redistribute load among the remaining representative experts \emph{without modifying any weights}.

The ``Subset Difference'' in the method's name refers to the mask construction: the set of experts pruned at level $L_2$ is the \emph{set difference} of the experts selected at two adjacent pruning thresholds, $\text{set}_{L_2} \setminus \text{set}_{L_1}$, deliberately skipping the top-10\% core tier.
This asymmetric removal targets the \emph{middle} of the redundancy spectrum—experts that are neither maximally representative nor maximally rare—which our analysis identifies as the source of functional overlap in Qwen3.5-35B-A3B.

%%%%%%%%%%%%%%%%%%%%%%%%%%%%%%%%%%%%%%%%%%%%%%%%%%%%%%%%%%%%

\section{Full Unique Expert Trajectory Figures}
\label{app:trajectories}

\begin{figure}[t]
\centering
\small \textbf{Unique Expert Count Trajectory of Gemma-4-27B-A4B} \\ [1.0ex]
\begin{subfigure}[b]{0.31\textwidth}
    \centering
    \includegraphics[width=\textwidth]{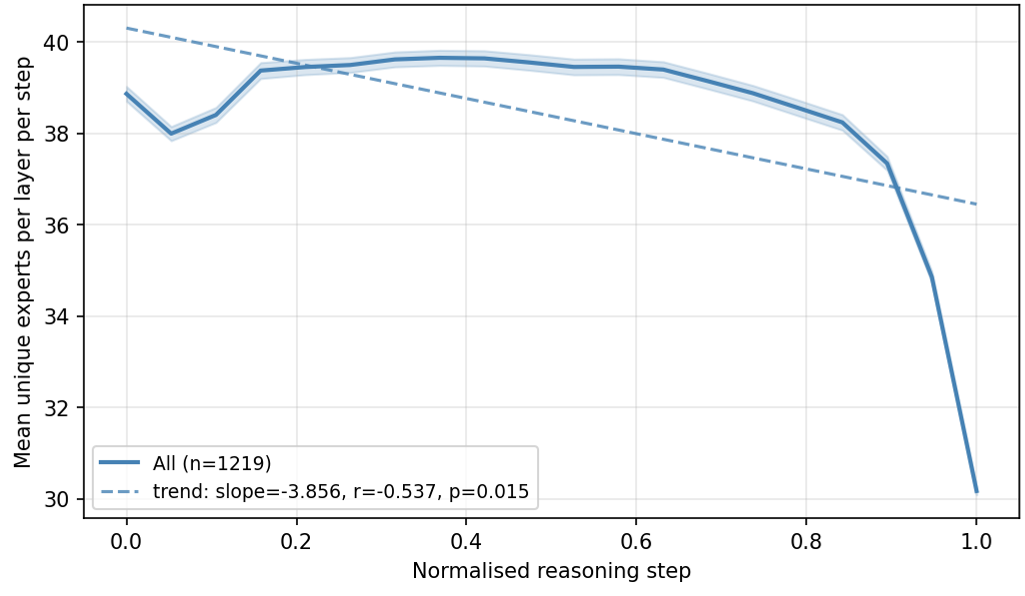}
    \caption{GSM8K}
\end{subfigure}
\hfill
\begin{subfigure}[b]{0.31\textwidth}
    \centering
    \includegraphics[width=\textwidth]{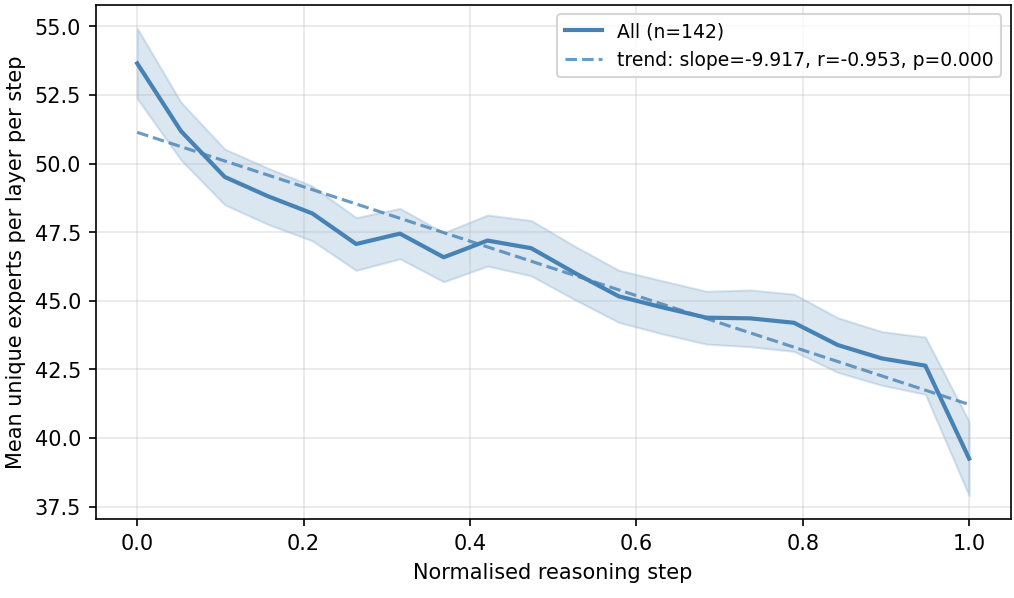}
    \caption{AQuA}
\end{subfigure}
\hfill
\begin{subfigure}[b]{0.31\textwidth}
    \centering
    \includegraphics[width=\textwidth]{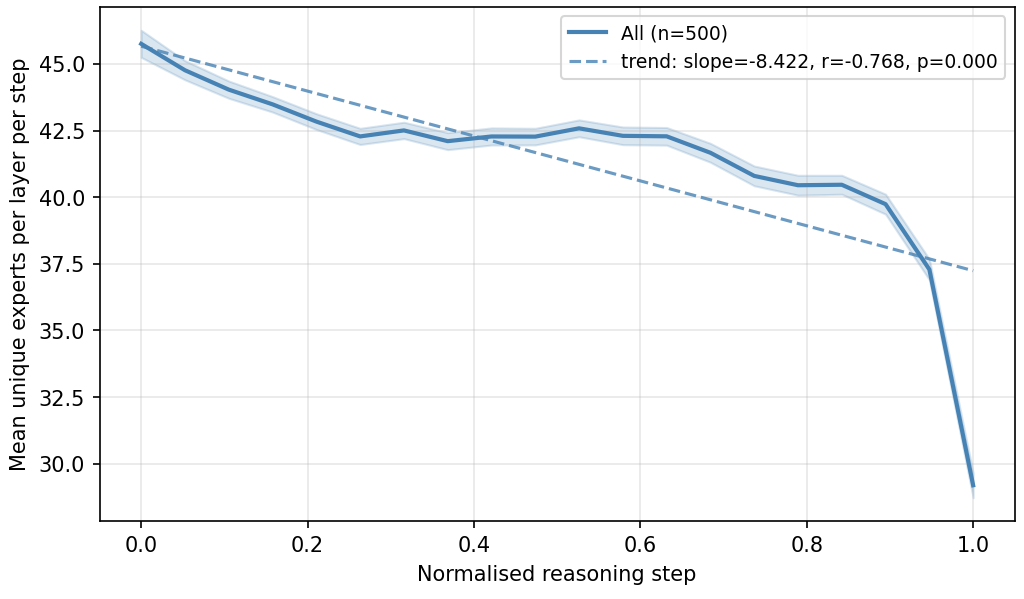}
    \caption{CompMath-MCQ}
\end{subfigure}
\vspace{1ex}
\caption{Unique expert count trajectory for \textbf{Gemma-4-27B-A4B} across three datasets (GSM8K, AQuA, and CompMath-MCQ); the MATH trajectory appears in Figure~\ref{fig:unique_expert_comparison}. All datasets show consistent monotonic decline.}
\label{fig:traj_gemma}
\end{figure}

\begin{figure}[t]
\centering
\small \textbf{Unique Expert Count Trajectory of Phi-3.5-MoE} \\ [1.0ex]
\begin{subfigure}[b]{0.31\textwidth}
    \centering
    \includegraphics[width=\textwidth]{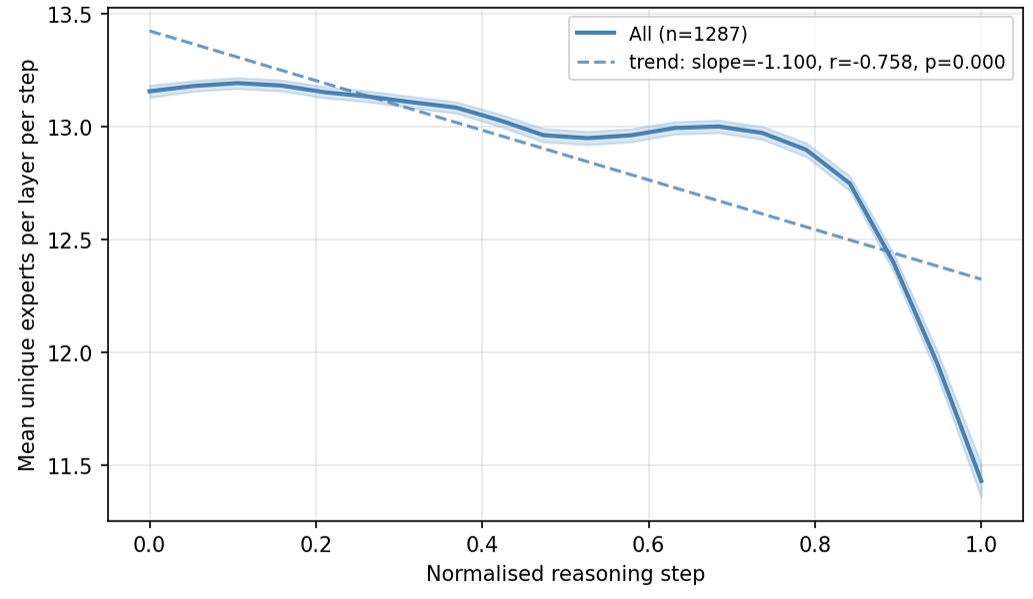}
    \caption{GSM8K}
\end{subfigure}
\hfill
\begin{subfigure}[b]{0.31\textwidth}
    \centering
    \includegraphics[width=\textwidth]{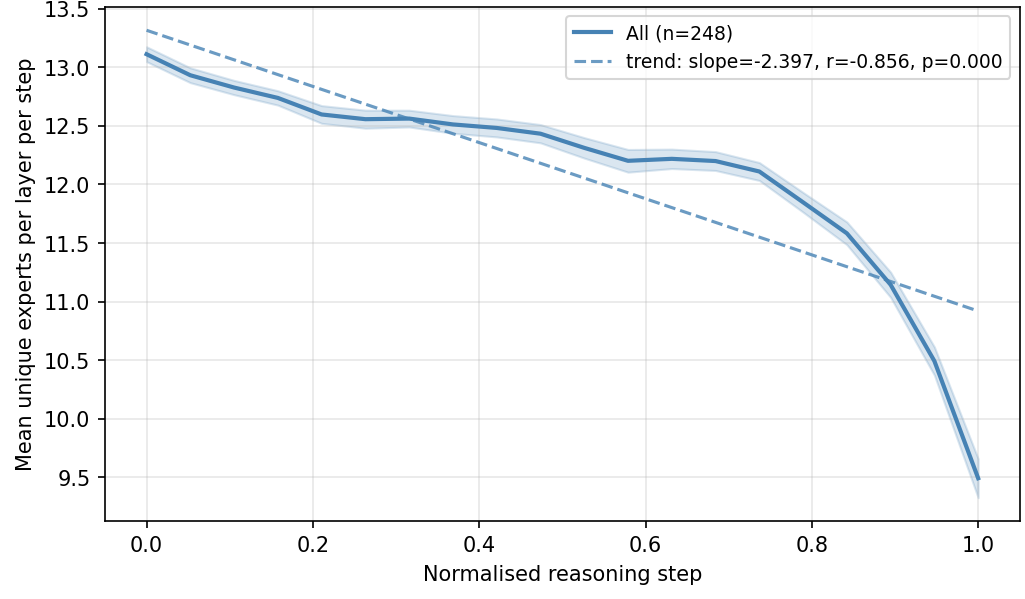}
    \caption{AQuA}
\end{subfigure}
\hfill
\begin{subfigure}[b]{0.31\textwidth}
    \centering
    \includegraphics[width=\textwidth]{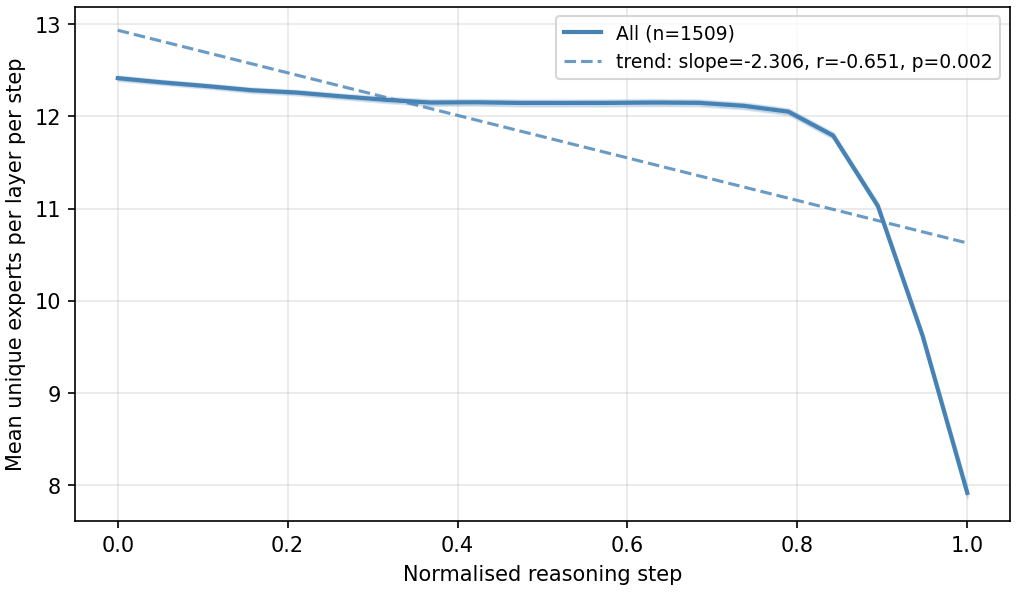}
    \caption{CompMath-MCQ}
\end{subfigure}
\vspace{1ex}
\caption{Unique expert count trajectory for \textbf{Phi-3.5-MoE} across three datasets (GSM8K, AQuA, and CompMath-MCQ); the MATH trajectory appears in Figure~\ref{fig:unique_expert_comparison}. The oscillating but overall declining pattern is consistent across both open-ended and MCQ tasks.}
\label{fig:traj_phi}
\end{figure}

\begin{figure}[t]
\centering
\small \textbf{Unique Expert Count Trajectory of Qwen3.5-35B-A3B} \\ [1.0ex]
\begin{subfigure}[b]{0.31\textwidth}
    \centering
    \includegraphics[width=\textwidth]{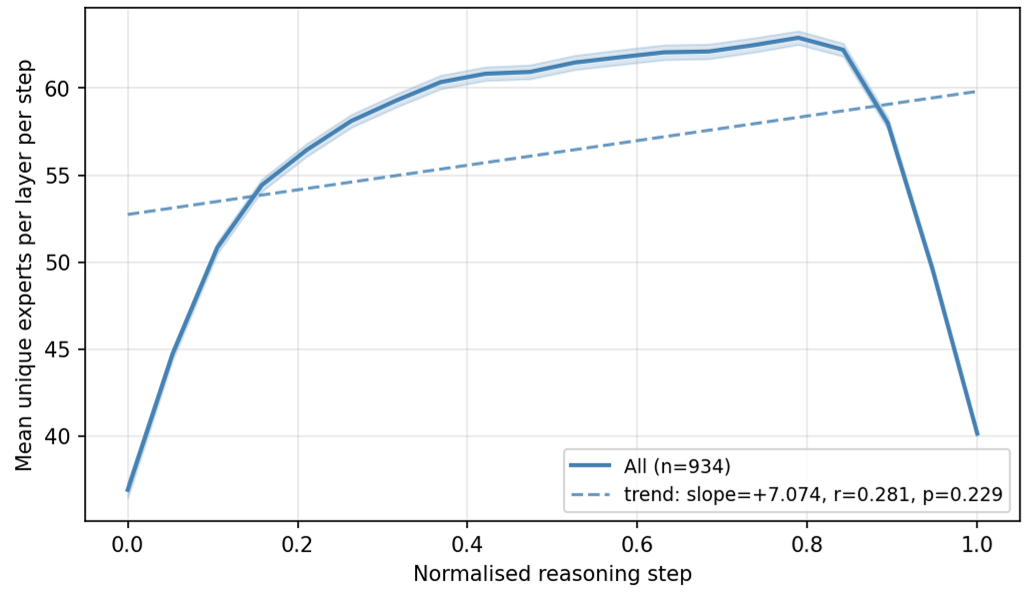}
    \caption{GSM8K}
\end{subfigure}
\hfill
\begin{subfigure}[b]{0.31\textwidth}
    \centering
    \includegraphics[width=\textwidth]{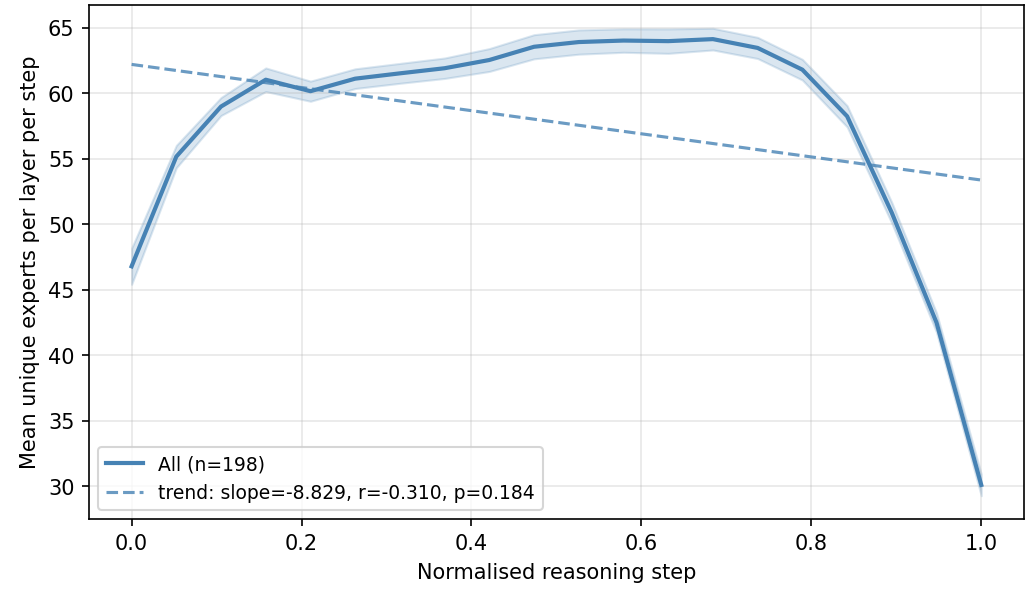}
    \caption{AQuA}
\end{subfigure}
\hfill
\begin{subfigure}[b]{0.31\textwidth}
    \centering
    \includegraphics[width=\textwidth]{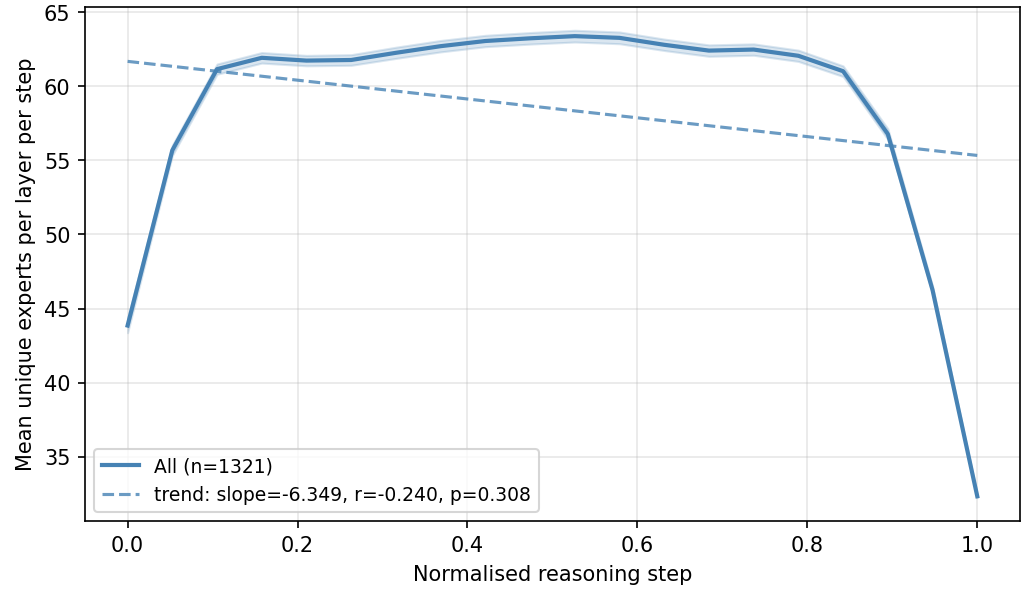}
    \caption{CompMath-MCQ}
\end{subfigure}
\vspace{1ex}
\caption{Unique expert count trajectory for \textbf{Qwen3.5-35B-A3B} across three datasets (GSM8K, AQuA, and CompMath-MCQ); the MATH trajectory appears in Figure~\ref{fig:unique_expert_comparison}. The inverted-U pattern (exploration then late collapse) is consistent across tasks, reflecting the model's divergent exploration strategy under fine-grained ($k/E=3.1\%$) routing.}
\label{fig:traj_qwen}
\end{figure}

Figures~\ref{fig:traj_gemma}--\ref{fig:traj_qwen} present the unique expert count trajectory for all three models across all four datasets. The $x$-axis is the normalized reasoning step position (0 = first step, 1 = last step); the $y$-axis is mean unique experts per sensitive layer per step; the dashed line is an OLS linear trend with slope and Pearson $r$ annotated.

%%%%%%%%%%%%%%%%%%%%%%%%%%%%%%%%%%%%%%%%%%%%%%%%%%%%%%%%%%%%

\section{Empirical Validation of Temporal Compression (Eq.~\ref{eq:temporal_decay})}
\label{app:temporal_decay}

We estimate $\hat{H}(O \mid \text{step}\,i)$ by partitioning each trace into ten equal-width normalized-position bins and computing the entropy of the empirical operation-type distribution within each bin.
Figure~\ref{fig:temporal_decay_fit} (main paper) plots the resulting $(\hat{H}, \bar{u}_i)$ pairs for GSM8K.

%%%%%%%%%%%%%%%%%%%%%%%%%%%%%%%%%%%%%%%%%%%%%%%%%%%%%%%%%%%%

\section{Expert Routing Specialization by Operation Type}
\label{app:specialization}

The Frequency-Diversity Law (Section~\ref{sec:huffman}) establishes that frequent operations receive shorter routing codes---fewer active experts per step.
A natural follow-up question is whether these codes are merely \emph{compact}, or also \emph{consistent}: do steps of the same operation type, drawn from entirely unrelated problems, activate the same expert subsets?

\paragraph{Code Consistency Across Problem Instances.}
Beyond the frequency-diversity correspondence, the ``short codes'' assigned to late-stage reasoning steps are not arbitrary: they encode the \emph{type of arithmetic operation} in a content-independent manner.
Steps performing division drawn from unrelated problems activate nearly identical expert subsets ($J$ up to $0.947$), whereas cross-operation pairs yield substantially lower overlap (Table~\ref{tab:specialization_examples}).
This \emph{routing fingerprint consistency} property is a direct implication of Huffman optimality: each semantic operation type $o \in \mathcal{O}$ is assigned a stable, reproducible routing fingerprint $\mathcal{E}^*(o)$ that is invariant to surface-level problem content.
Crucially, this property holds across all three architectures, indicating that code consistency is a structural consequence of routing optimisation rather than any architecture-specific inductive bias.

To quantify this, we classify each thought step by semantic operation type (add/subtract, multiply/divide, problem setup, restatement, other) using a lightweight keyword classifier whose trigger words are listed in Table~\ref{tab:op_keywords}. Routing overlap is measured by the pairwise Jaccard similarity between expert sets:

\begin{table}[ht]
\caption{Keyword triggers for the operation-type classifier. A step is assigned to the first matching category in the order shown; steps matching none of the listed keywords are labelled \texttt{other}.}
\label{tab:op_keywords}
\centering
\small
\begin{tabular}{ll}
\toprule
Category & Trigger keywords \\
\midrule
\texttt{setup}
  & \textit{let, let's, suppose, given that, we know, we have,} \\
  & \textit{step 1, first, define, note that} \\
\midrule
\texttt{add/subtract}
  & \textit{add, subtract, sum, total, plus, minus, difference,} \\
  & \textit{increase, decrease, more than, less than,} \texttt{+}\textit{,} \texttt{-} \\
\midrule
\texttt{multiply/divide}
  & \textit{multiply, times, product,} $\times$\textit{, per, each, rate,} \\
  & \textit{fraction, divide, divided, ratio, percent,} \texttt{\%} \\
\bottomrule
\end{tabular}
\end{table}

\begin{equation}
    J\!\left(\mathcal{E}_a^{(l)},\, \mathcal{E}_b^{(l)}\right)
    = \frac{\left|\mathcal{E}_a^{(l)} \cap \mathcal{E}_b^{(l)}\right|}
           {\left|\mathcal{E}_a^{(l)} \cup \mathcal{E}_b^{(l)}\right|}.
    \label{eq:jaccard}
\end{equation}
We then compute for each model--dataset pair:
\begin{itemize}[leftmargin=1.2em, itemsep=0.2em]
    \item $\bar{J}_{\text{intra}}$: mean pairwise Jaccard between steps of the \emph{same} operation type drawn from \emph{different} problems.
    \item $\bar{J}_{\text{inter}}$: mean pairwise Jaccard between arithmetic and non-arithmetic (explain/restate) steps.
\end{itemize}
A positive $\Delta = \bar{J}_{\text{intra}} - \bar{J}_{\text{inter}}$ confirms that routing encodes operation type beyond problem identity.

Table~\ref{tab:specialization} shows $\Delta > 0$ in all 12 model--dataset combinations, with Mann--Whitney $p \le 10^{-163}$ in every case.
The absolute $\bar{J}$ values vary across models because they reflect the expert sparsity ratio $k/E$ (Phi: $2/16=12.5\%$; Gemma: $8/128=6.25\%$; Qwen: $8/256=3.1\%$)---higher sparsity means higher expected baseline overlap---but the \emph{relative} gap $\Delta \in [+0.02, +0.10]$ is consistent across all architectures.

\begin{table}[ht]
\caption{%
    Expert routing specialization by arithmetic operation type (Mann--Whitney U test).
    $\Delta > 0$ in all 12 cases confirms that the router encodes \emph{what computation is performed},
    independent of problem content and model architecture.
    Absolute $\bar{J}$ values scale with sparsity ratio $k/E$ (shown as $E$ column).
}
\label{tab:specialization}
\centering
\small
\begin{tabular}{llcccrc}
\toprule
Model & Dataset & $\bar{J}_{\text{intra}}$ & $\bar{J}_{\text{inter}}$ & $\Delta$ & $p$-value & $E$ \\
\midrule
\multirow{4}{*}{Gemma-4-27B-A4B}
  & GSM8K        & 0.526 & 0.475 & \textbf{+0.051} & $<10^{-300}$ & 128 \\
  & MATH         & 0.543 & 0.502 & \textbf{+0.040} & $<10^{-300}$ & 128 \\
  & AQuA         & 0.528 & 0.475 & \textbf{+0.053} & $<10^{-300}$ & 128 \\
  & CompMath-MCQ & 0.530 & 0.458 & \textbf{+0.072} & $<10^{-300}$ & 128 \\
\midrule
\multirow{4}{*}{Phi-3.5-MoE}
  & GSM8K        & 0.809 & 0.755 & \textbf{+0.055} & $<10^{-300}$ & 16  \\
  & MATH         & 0.719 & 0.701 & \textbf{+0.018} & $<10^{-163}$ & 16  \\
  & AQuA         & 0.775 & 0.727 & \textbf{+0.048} & $<10^{-300}$ & 16  \\
  & CompMath-MCQ & 0.785 & 0.689 & \textbf{+0.095} & $<10^{-300}$ & 16  \\
\midrule
\multirow{4}{*}{Qwen3.5-35B-A3B}
  & GSM8K        & 0.331 & 0.259 & \textbf{+0.072} & $<10^{-300}$ & 256 \\
  & MATH         & 0.309 & 0.268 & \textbf{+0.042} & $<10^{-300}$ & 256 \\
  & AQuA         & 0.315 & 0.236 & \textbf{+0.079} & $<10^{-300}$ & 256 \\
  & CompMath-MCQ & 0.342 & 0.251 & \textbf{+0.090} & $<10^{-300}$ & 256 \\
\bottomrule
\end{tabular}
\end{table}

\paragraph{Qualitative illustration.}
Table~\ref{tab:specialization_examples} shows the highest-Jaccard routing matches found between steps from \emph{entirely different problems} performing the same arithmetic operation.
Despite sharing no numbers, variables, or narrative context, these step pairs activate nearly identical expert subsets---with Jaccard up to $J=0.947$ (Phi-3.5-MoE on GSM8K).
Two steps that both say ``divide both sides by $c$ to solve for $x$'' (different $c$, different equations, different problems) route through virtually the same expert combination: the router has learned a stable, problem-agnostic ``division'' code---exactly what the Huffman analogy predicts.

\begin{table}[ht]
\caption{%
    Highest-Jaccard cross-problem routing matches for same-type arithmetic steps.
    Step pairs are drawn from unrelated questions; text is truncated at 60 characters.
    High Jaccard across unrelated problems confirms operation-type encoding is content-independent.
}
\label{tab:specialization_examples}
\centering
\small
\renewcommand{\arraystretch}{1.25}
\begin{tabular}{p{2.8cm} l c p{6.8cm}}
\toprule
Model & Dataset & $J$ & Step pair (truncated) \\
\midrule
Gemma-4-27B-A4B & MATH & 0.866
  & \textit{``Divide both sides by 2 to solve for $x$.''} \\
  & & & \quad vs.\ \textit{``Divide both sides by 3 to solve for $n$.''} \\
\midrule
Phi-3.5-MoE & GSM8K & 0.947
  & \textit{``Step 4: Calculate the total amount of time Melissa spends on\dots''} \\
  & & & \quad vs.\ \textit{``Step 4: Calculate the overall probability that Marcus won't\dots''} \\
\midrule
% Phi-3.5-MoE & COMPMATH & 0.909
%   & \textit{``Step 4: For matrix A, the characteristic equation is $(2-\lambda)$\dots''} \\
%   & & & \quad vs.\ \textit{``Step 4: Find the eigenvalues of matrix A\dots''} \\
% \midrule
Qwen3.5-35B-A3B & COMPMATH & 0.658
  & \textit{``The determinant is calculated as 3 multiplied by 5 minus\dots''} \\
  & & & \quad vs.\ \textit{``The determinant is computed as 2 times 14 minus $-5$ times $-5$.''} \\
\bottomrule
\end{tabular}
\end{table}

%%%%%%%%%%%%%%%%%%%%%%%%%%%%%%%%%%%%%%%%%%%%%%%%%%%%%%%%%%%%

\section{Quantitative Huffman Validation: $\bar{u}^{(o)} \propto {-}\log p(o)$}
\label{app:huffman_validation}

This appendix reports the full numerical results for the quantitative Huffman test (Eq.~\ref{eq:huffman} in the main paper) and explains why Qwen3.5-35B-A3B is excluded from the claim.

\paragraph{Setup.}
We classify each thought step into one of four types using a keyword classifier:
\emph{add/subtract} (keywords: sum, plus, minus, difference, \ldots),
\emph{multiply/divide} (keywords: times, product, percent, ratio, \ldots),
\emph{setup} (keywords: let, given, suppose, define, \ldots), and
\emph{other}.
For each model we compute the empirical frequency $p(o)$ and the mean unique expert count $\bar{u}^{(o)}$ across all correct traces and all four datasets.

In total this yields \textbf{756 correct traces} (6{,}688 classified steps, excluding restate) for Gemma-4-27B-A4B, and \textbf{618 correct traces} (4{,}807 classified steps) for Phi-3.5-MoE.
Each scatter point in Figure~\ref{fig:huffman_scatter} therefore aggregates hundreds to thousands of individual step observations, providing strong statistical backing for the Spearman $\rho = 1.00$ result.

\paragraph{Gemma-4-27B-A4B and Phi-3.5-MoE: perfect Huffman rank order.}
Table~\ref{tab:huffman_validation} reports the per-type statistics for the two models that satisfy Eq.~\ref{eq:huffman}.
Both models achieve \textbf{Spearman $\rho = 1.00$} ($p < 0.005$) and Pearson $r > 0.97$ ($p < 0.04$).
The rank order $\bar{u}(\text{other}) < \bar{u}(\text{add/sub}) < \bar{u}(\text{mult/div}) < \bar{u}(\text{setup})$ mirrors frequency rank in reverse across both architectures, confirming the Huffman prediction is not model-specific.

\begin{table}[ht]
\caption{%
    Per-type statistics for the quantitative Huffman test (aggregated over all four datasets, correct traces only).
    $n$ = number of classified steps (restate excluded); $p(o)$ = empirical frequency; $\bar{u}^{(o)}$ = mean unique experts per layer.
    Gemma: 756 questions, 6{,}688 steps total; Phi: 618 questions, 4{,}807 steps total; Qwen: shown for reference only (does not satisfy Eq.~\ref{eq:huffman}).
}
\label{tab:huffman_validation}
\centering
\small
\begin{tabular}{llrrrr}
\toprule
Model & Operation type & $n$ & $p(o)$ & $-\log p(o)$ & $\bar{u}^{(o)}$ \\
\midrule
\multirow{4}{*}{Gemma-4-27B-A4B}
  & Other          & 2840 & 0.374 & 0.98 & 37.3 \\
  & Add/Subtract   & 1736 & 0.229 & 1.47 & 40.9 \\
  & Mult/Divide    & 1439 & 0.190 & 1.66 & 41.4 \\
  & Setup          &  673 & 0.089 & 2.42 & 53.6 \\
\midrule
\multirow{4}{*}{Phi-3.5-MoE}
  & Other          & 1696 & 0.304 & 1.19 & 11.6 \\
  & Add/Subtract   & 1132 & 0.203 & 1.59 & 12.2 \\
  & Setup          & 1047 & 0.188 & 1.67 & 12.5 \\
  & Mult/Divide    &  932 & 0.167 & 1.79 & 12.6 \\
\midrule
\multirow{4}{*}{Qwen3.5-35B-A3B}
  & Other          & 3456 & 0.412 & 0.89 & 58.3 \\
  & Add/Subtract   & 1756 & 0.209 & 1.56 & 66.9 \\
  & Mult/Divide    & 1437 & 0.171 & 1.76 & 62.2 \\
  & Setup          &  799 & 0.095 & 2.35 & 45.3 \\
\bottomrule
\end{tabular}
\end{table}

\paragraph{Why Qwen3.5-35B-A3B violates Eq.~\ref{eq:huffman}.}
With $E{=}256$ experts and $k{=}8$ ($k/E = 3.1\%$), the expected pairwise overlap between any two independently-sampled steps is nearly zero.
Table~\ref{tab:huffman_validation} shows the resulting disorder: \emph{setup} steps (the rarest type, $p{=}0.095$) have the \emph{lowest} $\bar{u}$ (45.3), while \emph{add/subtract} steps (more common, $p{=}0.209$) have the \emph{highest} $\bar{u}$ (66.9).
As explained in Section~\ref{sec:qwen_pruning}, the root cause is Qwen's inverted-U trajectory: expert diversity peaks mid-trace and collapses toward the end, so the position of each operation type along this curve---not its frequency---determines $\bar{u}^{(o)}$.
Qwen therefore exhibits the qualitative temporal compression pattern (inverted-U exploration followed by late-stage collapse) but does not satisfy the quantitative frequency--diversity relationship of Eq.~\ref{eq:huffman}.

%%%%%%%%%%%%%%%%%%%%%%%%%%%%%%%%%%%%%%%%%%%%%%%%%%%%%%%%%%%%

\section{Limitations}
\label{app:limitations}

\emph{(1) Domain scope.}
All experiments are restricted to mathematical reasoning benchmarks (GSM8K \citep{cobbe2021training}, MATH \citep{hendrycks2021measuring}, AQuA \citep{ling2017program}, CompMath-MCQ \citep{lo2025closer}).
Whether the Frequency-Diversity Law generalizes to code generation, factual question-answering, or open-ended generation remains an open question.

\emph{(2) Operation-type classifier.}
We classify reasoning steps via a keyword-based heuristic into four types.
With only $|\mathcal{O}|{=}4$ categories, the Pearson correlation in the Huffman scatter is estimated from four points per model; the result is not statistically significant for Qwen's baseline ($p{=}0.37$) and only marginally so for the two Huffman-compliant models ($p{<}0.05$).
A richer, model-assisted classifier producing more fine-grained operation types would increase statistical power.

\emph{(3) Benchmark scope for Qwen pruning.}
The pruning evaluation (Table~\ref{tab:qwen_pruning_acc}) is conducted on GSM8K; accuracy on the full 1{,}319-question benchmark and on the other three datasets (MATH, AQuA, CompMath-MCQ) under each pruning level remains to be evaluated.

% \emph{(4) Static pruning masks.}
% Subset Difference Pruning computes co-activation profiles offline and applies a fixed mask at inference time.
% The pruning is not adaptive to the current input or reasoning context, which may limit performance on out-of-distribution problems.

\emph{(4) Huffman $r$ as a training-time signal.}
Our results suggest that $r$ could be monitored during training as an early indicator of redundancy formation: a persistent negative $r$ would signal that load-balancing regularization is creating functional near-duplicates faster than the model develops meaningful expert specialization.
Whether online $r$ monitoring can guide adaptive regularization schedules (or serve as a stopping criterion for load-balancing loss) remains an open question for future work.

%%%%%%%%%%%%%%%%%%%%%%%%%%%%%%%%%%%%%%%%%%%%%%%%%%%%%%%%%%%%

%%%%%%%%%%%%%%%%%%%%%%%%%%%%%%%%%%%%%%%%%%%%%%%%%%%%%%%%%%%%%%%%%%%%%%%%%%%%%%%
%%%%%%%%%%%%%%%%%%%%%%%%%%%%%%%%%%%%%%%%%%%%%%%%%%%%%%%%%%%%%%%%%%%%%%%%%%%%%%%

\end{document}